\documentclass[11pt]{article}

\usepackage[preprint]{acl}

\usepackage{times}
\usepackage{latexsym}
\usepackage{tcolorbox}
\tcbuselibrary{breakable, skins}
\usepackage{tabularx}
\usepackage{booktabs}
\usepackage{enumitem}
\usepackage{xurl}
\usepackage{hyperref}
\usepackage[T1]{fontenc}
\usepackage{subcaption}
\usepackage[utf8]{inputenc}

\usepackage{microtype}

\usepackage{inconsolata}

\usepackage{graphicx}
\usepackage{bm}
\usepackage{multirow}
\usepackage{tcolorbox}
\usepackage[table]{xcolor}
\usepackage{booktabs}
\usepackage{amsmath}
\usepackage{subcaption}
\usepackage{pifont}   
\usepackage{fontawesome}
\usepackage{amssymb}  
\usepackage[switch]{lineno} 

\usepackage{xcolor, soul,todonotes} 
\definecolor{kmycolor}{rgb}{0.858, 0.188, 0.478}

\newcommand{\rr}[0]{}

\newcommand{\ds}[0]{\textsc{PolBench}}
\newcommand{\ours}[0]{\textsc{GitSearch}}

\title{\textsc{GitSearch}: Enhancing Community Notes Generation\\ with Gap-Informed Targeted Search}

\author{Sahajpreet Singh\textsuperscript{1} ~~~ Kokil Jaidka\textsuperscript{1,$\dagger$, \faEnvelopeO} ~~~ Min-Yen Kan\textsuperscript{1,$\dagger$, \faEnvelopeO} ~~~ Preslav Nakov\textsuperscript{2,$\dagger$} \\
$^1$National University of Singapore\\
$^2$Mohamed bin Zayed University of Artificial Intelligence \\
\textsuperscript{\faEnvelopeO~}Corresponding author(s): \texttt{\{jaidka, knmnyn\}@nus.edu.sg}
}

\begin{document}
\maketitle
\begin{abstract}
\rr{
Community Notes enable scalable, community-based fact-checking, but automated generation faces a coverage--quality trade-off. Note-dependent methods (e.g., summarization-based Supernotes) cannot operate without prior crowd input, while web-search LLMs achieve broad coverage but often retrieve evidence misaligned with what raters need. We introduce \ours{} (Gap-Informed Targeted Search), which treats missing, ambiguous, or disputed information as first-class retrieval targets. It infers information gaps from a post and optional incomplete notes, searches specifically to resolve them, and synthesizes an evidence-grounded Community Note. To evaluate our framework, we introduce \ds{}, a new dataset of 78,698 U.S.\ political tweets with Community Notes. Keeping single-call search inference budget fixed, \ours{} improves human-rated helpfulness by 0.43 over unconditioned web search and wins 59\% of pairwise comparisons. It generates notes for 99\% of test cases, including the 46\% where summarization methods fail. Human evaluators prefer \ours{} to human-written helpful notes in 69\% of comparisons, primarily due to better context coverage. These findings show that how search is organized matters more than information access alone.
}
\end{abstract}

\section{Introduction}
\label{sec:intro}

{\let\thefootnote\relax\footnotetext{$^\dagger$Equal supervisory contribution.}}

{\let\thefootnote\relax\footnotetext{
\textbf{Data \& Code:} \href{https://github.com/sahajps/GitSearch}{\texttt{github.com/sahajps/GitSearch}}}}

{\let\thefootnote\relax\footnotetext{
\textcolor{red}{\textbf{Disclaimer:} This paper contains examples of misleading and harmful content used solely for illustration purposes.}}}

\begin{figure}[!t]
\centering
\includegraphics[width=\columnwidth]{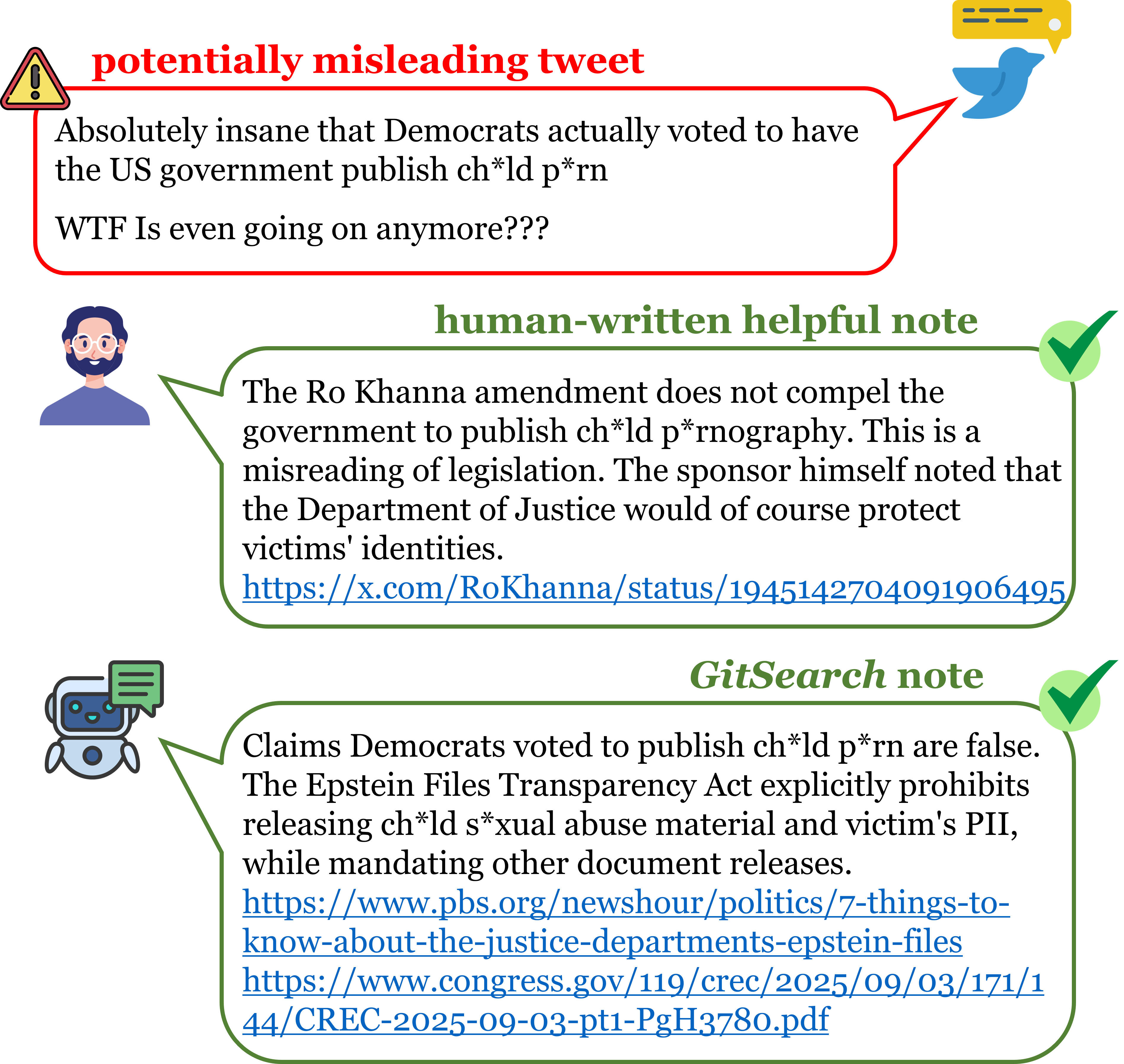}
\caption{Example of human-written and \ours{}-generated notes addressing misleading tweet.}
\label{fig:gitsearch_intro}
\vspace{-\baselineskip}
\end{figure}

Moderating misinformation on social media is critical for effective governance and intervention in online discourse \cite{lazer2018science}. Distributed, user-driven moderation can complement centralized fact-checking by placing contextual information directly alongside disputed posts. \rr{For instance, Community Notes (X’s crowdsourced fact-checking feature) enables volunteer contributors to provide contextual notes on potentially misleading posts. These notes are publicly displayed only when contributors with diverse perspectives collectively rate them as helpful \cite{slaughter2025community}.} Despite their effectiveness, Community Notes face persistent structural challenges. First, \textbf{scale and coverage} remain limited: generating timely notes requires sustained human participation; many misleading posts, especially emerging or event-driven claims, remain uncovered for extended periods \cite{wu2025beyond}. Our analysis shows that approximately 92\% of notes do not receive sufficient \rr{diverse} ratings to be classified as either helpful or unhelpful, resulting in a vast majority labeled as needing more ratings. Second, \textbf{note quality is uneven}. Even when notes exist, they vary substantially in completeness, sourcing quality, neutrality, and adherence to platform recommendations \cite{de2025supernotes}.  Moreover, we find that the share of notes ultimately deemed unhelpful exceeds that of helpful notes (4.19\% vs. 3.57\%). Third, existing systems make limited use of the \textbf{information gaps} revealed by notes. Partially written, conflicting, or incomplete notes often signal \textit{what} contextual or evidentiary information human raters perceive as missing; yet, prior work \cite{de2025supernotes,wu2025beyond} has rarely operationalized these signals to guide automated moderation. Crucially, these gaps represent not merely missing facts, but misalignments with rater expectations. In Community Notes, helpfulness depends on how raters perceive a note's relevance, completeness, and evidentiary support for a specific post \cite{solovev2025references}. Notes may be rated unhelpful despite factual accuracy if they omit salient context, address the wrong claim, lack substantiation that raters expect, or are due to biases in annotations \cite{truong2025community,kuuse2025crowdsourced}.

\rr{Our instrument for testing information gaps phenomena is} \ours\ (\textbf{G}ap-\textbf{I}nformed-\textbf{T}argeted \textbf{Search}), an AI-based dynamic real-time Community Note generation framework treating perceived quality gaps as first-class signals. \ours\ operates in both ``cold start'' (no existing notes) and ``warm start'' (notes labeled \texttt{NEEDS\_MORE\_RATINGS} or \texttt{NMR} tag) scenarios via a three-stage pipeline to synthesize gap-targeted, platform-compliant notes while maintaining neutrality and sourcing standards.
For our experiments, we have curated a dataset of 78,698 U.S. political tweets paired with human-authored Community Notes spanning events before and after major LLMs' knowledge cutoffs, which we refer to as \ds{}. \ours{} produces notes for 99\% of test cases, including the 46\% where summarization baselines cannot operate due to the lack of existing notes. It also surpasses human-authored helpful notes in head-to-head evaluation (69\% win rate; helpfulness: 3.87 vs.\ 3.36). Critically, when controlling for model and \rr{single-call search budget}, \ours{} outperforms generic web-search agents (59\% win rate), demonstrating that \textit{retrieval structure}, not model capability, is the decisive factor in note quality.

\rr{We make the following contributions:}
\begin{itemize}[leftmargin=*]
    \vspace{-2mm}
    \item \rr{We introduce \ours{}, a framework that identifies missing information, searches for relevant evidence, and generates grounded notes, even in cold-start scenarios.}
    \vspace{-2mm}
    \item \rr{We show that searching for specific information gaps produces better notes than general web search under the same search budget. Our approach achieves 99\% coverage, compared to 54\% for the strongest note-dependent baseline, without sacrificing note quality.}
    \vspace{-2mm}
    \item \rr{We introduce \ds{}, a dataset of 78,698 U.S. political tweets and their associated Community Notes, which we release to support future research on automated note generation.}
\end{itemize}

\section{Related Work}
\label{sec:rel_work}
\paragraph{Misinformation and Community Moderation.}
Misinformation on social media can affect political discourse, civic engagement, and public health \cite{lazer2018science,mcdougall2019media}. Prior studies have also shown its effects on people's beliefs, even when exposure is limited \cite{tokita2021polarized,kim2022consequences}. Community-based moderation offers an alternative to centralized fact-checking by allowing users to provide context for potentially misleading posts \cite{micallef2020role,martel2024crowds,pfander2025spotting}. However, these systems face practical limitations. Community Notes take a median of 17 hours to reach consensus \cite{wu2025beyond}, and most submitted notes never receive sufficient ratings \cite{de2025supernotes}. Moreover, contributors have varying levels of fact-checking expertise, and Community Notes often rely on professional fact-checking sources \cite{borenstein2025can}.

\paragraph{AI-assisted Community Notes Generation.}
Previous studies have explored using LLMs to debunk misinformation \cite{kim2024can,zhou2024correcting,singh2025limitations}. More recent work focuses on generating or improving Community Notes. Supernotes \cite{de2025supernotes} summarizes existing crowd-written notes but cannot operate without prior contributions. CrowdNotes+ \cite{wu2025beyond} retrieves supporting evidence using URLs identified by human contributors, although web retrieval can be limited by restrictions imposed by news publishers \cite{fletcher2024how}. Commenotes \cite{zhang2025commenotes} uses post comments to generate notes, while \citet{mohammadi2025ai} investigate how AI feedback can improve human-written notes. Related work by \citet{li2024re} uses iterative retrieval-augmented reasoning for fake news detection, but requires multiple retrieval steps. Importantly, these approaches either depend on existing human contributions or require additional retrieval steps. In contrast, \ours{} identifies missing information from the post and any available notes, and uses these gaps to guide a single web search. This allows us to generate notes even when no prior Community Notes exist, while keeping retrieval targeted.

\section{The \ours{} Framework}
\label{sec:framework}
We investigate whether decomposing Community Note generation into structured sub-tasks (i.e.,~gap identification, targeted retrieval, and constrained synthesis) can yield more helpful notes than end-to-end generation. 

\begin{figure}[th]
    \centering
    \includegraphics[width=\linewidth]{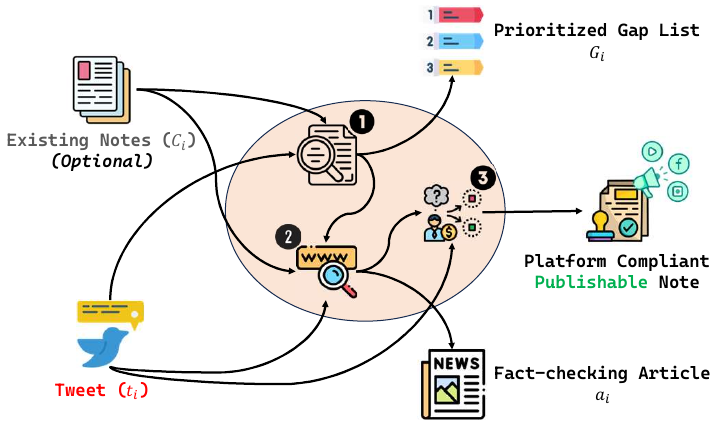}
    \caption{Overview of our three-phase (in black numbering) \ours{} framework. 
    }
    \label{fig:gitsearch_framework}
    \vspace{-\baselineskip}
\end{figure}

\subsection*{Problem Formulation}
Consider the dataset $\mathcal{D}=\{(t_i, n_i, \mathcal{C}_i)\}$, where $t_i \in \mathcal{T}$ represents the $i$-th target post (tweet), and $n_i \in \mathcal{N}$ represents a ``helpful'' Community Note associated with $t_i$. The set $\mathcal{C}_i$ contains candidate notes (if any) currently labeled \texttt{NMR}, representing incomplete or non-consensus crowd efforts.

\rr{Our objective is to generate a Community Note $\hat{n}_i = f_\theta(t_i,\mathcal{C}_i)$ that accurately addresses the claims in $t_i$ using relevant evidence.} This alignment is evaluated along three axes: \textit{factual correctness} (validity of claims), \textit{contextual completeness} (coverage of necessary background), and \textit{utility} (alignment with community helpfulness guidelines). \ours{} (Figure~\ref{fig:gitsearch_framework}) maps gap identification to evidence retrieval and constrained synthesis, addressing the coverage and hallucination limitations inherent in standard end-to-end generation pipelines.

\subsection*{Phase I: Gap Detection \& Prioritization}
\rr{Before retrieving anything, \ours{} first determines \textit{what is missing}. These deficits are inferred from two sources: the tweet's own claims, and the incomplete or conflicting \texttt{NMR} notes that reveal what raters found insufficient. Following annotation schemas for claim detection \cite{konstantinovskiy2021toward,brennen2020types}, we operationalize six types of information gaps that recur in Community Notes and serve as the structured signal conditioning retrieval: \textit{unsubstantiated claim}, \textit{contradiction}, \textit{vague reference}, \textit{missing context}, \textit{source verification}, and \textit{missing coverage} (definitions in Appendix~\ref{app:def_gaps}). The categories were derived through inductive content analysis \cite{hsieh2005three} of (tweet, \texttt{NMR} note) pairs from \ds{} (Section~\ref{sec:dataset}), iteratively refined until stable, covering both deficiencies within $t_i$ and inconsistencies across $\mathcal{C}_i$. We assess the taxonomy's empirical coverage in Appendix~\ref{app:gap_completeness}, finding that it captures the information deficits needed for targeted search in 82\% of sampled cases and its downstream effect on note quality in Section~\ref{sec:abl_study}. A separate manual audit finds that Gemini-2.5-Flash identifies accurate, well-grounded gaps in 85.4\% of sampled cases, with substantial inter-annotator agreement (Krippendorff’s $\alpha=0.79$; see Appendix~\ref{app:gap_accuracy} for detail).}

\paragraph{Gap Detection using LLM:} Gap detection is operationalized using an LLM prompted zero-shot with the full taxonomy definitions and exemplars. For each input tuple $(t_i, \mathcal{C}_i)$, the model returns a structured JSON list of gaps, each containing:
(a)~a gap type drawn from the taxonomy, (b)~a one-to-two sentence description, (c)~a suggested targeted retrieval query, and (d)~a severity score $s_g \in \{1, \ldots, 5\}$ on a Likert scale. Severity quantifies the gap's impact on the overall veracity of the claim. In zero-shot scenarios where $\mathcal{C}_i = \emptyset$, all unverified assertions in $t_i$ are treated as high-priority gaps by default. Removing this phase and passing $t_i$ directly to a web-search LLM corresponds to the Web-Agents baselines (Section \ref{sec:exp_setup}), isolating the contribution of structured gap detection.

For instance, given the tweet \textit{``Democrats actually voted to have the US government publish ch*ld p*rn''} with five conflicting \texttt{NMR} notes, Phase~I identifies a \textit{vague reference} gap (the unspecified legislation), a \textit{contradiction} (notes disagreeing on whether existing law would prevent publication), and an \textit{unsubstantiated claim} (the voting record assertion), each generating a targeted retrieval query for Phase~II (Appendix~\ref{app:intermediate_outputs}, Example~1).

\subsection*{Phase II: Targeted Search}
This phase executes a targeted search strategy, leveraging the prioritized gap set $\mathcal{G}_i$ along with the $(t_i, \mathcal{C}_i)$ pair to steer a web-search-enabled LLM in a single inference call. The model is explicitly instructed to treat each gap as a discrete search objective, retrieve evidence from credible and neutral sources, and synthesize findings into a coherent fact-checking article $a_i$. The query is conditioned specifically on high-priority gaps.

To ensure fair comparison, all Web-Agents baselines operate under the same single-call inference budget. One model invocation is performed with web-search enabled, using the same underlying LLM. The only difference is that \ours{} conditions the search on the structured gap set $\mathcal{G}_i$, whereas Web-Agents receive the raw tweet. This isolates retrieval \textit{structure} as the experimental variable, controlling for model capability and compute.

Continuing the example above, the \textit{vague reference} gap resolves to the \textit{``Epstein Files Transparency Act''}, which Phase~II retrieves alongside its explicit prohibition on releasing abuse material and the relevant roll-call record---evidence that directly addresses all three high-priority gaps in a single retrieval call.

\subsection*{Phase III: Constrained Note Synthesis}
The final phase distills the comprehensive evidence from Phase~II into a platform-compliant Community Note. A reasoning-capable LLM maps the original tweet $t_i$ and the evidence article $a_i$ to a final note $\hat{n}_i$, prohibiting reliance on parametric knowledge not present in the retrieved article.
\rr{Grounding is enforced by disabling web search during synthesis, instructing the model to omit claims unsupported by the retrieved article, and requiring all cited URLs to appear verbatim in that article. The full synthesis prompt is in Appendix \ref{app:prompt_temp}.}

The final candidate note is a raw text string with embedded URLs. Qualitative examples of all intermediate outputs, from gap detection through synthesis, are shown in Appendix~\ref{app:intermediate_outputs}. 

\section{Experimental Setup}
\label{sec:exp_setup}

\subsection{Baselines}
We design comparisons to isolate each phase's contribution:\\
\textit{(i)~Note summarization:}
Supernotes-Lite~\cite{de2025supernotes} summarizes \texttt{NMR} notes with helpfulness signals (fraction of helpful ratings) in-context, without web retrieval or an assessor component. This tests whether aggregating crowd efforts suffices, but is limited to the 54\% test cases with existing notes.\\
\textit{(ii)~Web-Agents:} Web-search LLMs~\cite{renault2025grok} receive raw tweets directly under the same single-call inference budget as Phase~II. This isolates the contribution of gap-driven retrieval.\\
\textit{(iii)~\ours{}:} Uses the full three-phase pipeline with a web-search-capable model as the backbone LLM, enabling direct comparison against the same models operating as Web-Agents.\\

We evaluate our setup across a representative set of models with varying capabilities (e.g., open-weight vs.\ proprietary, instruction-tuned vs.\ reasoning, with vs.\ without web search tool). Details on the models used and the web-agent post-processing are given in Appendices~\ref{app:model_details} and~\ref{app:post_processing_output}, respectively.

\subsection{Evaluation}
\rr{We evaluate \ours{} along three axes: automatic evaluation measures, reference-conditioned LLM-as-a-judge scores used diagnostically, and human evaluation as our primary measure. Human-written \texttt{CURRENTLY\_RATED\_HELPFUL (CRH)} notes serve as reference for similarity and LLM-based evaluation. However, they do not provide an upper bound on note quality, since \texttt{CRH} status can reflect stylistic fluency rather than usefulness \cite{wu2025beyond,xing2026communitynotes}. We use human evaluation, which does not rely on these references, as our primary measure.}

\paragraph{Similarity Scores \& Statistics:}
We assess content similarity using \emph{ROUGE-L} \cite{lin2004rouge} (lexical overlap), \emph{BERTScore} \cite{zhang2019bertscore} (semantic similarity), and \emph{URL Recall} (overlap with URLs cited in the reference note). We also report \emph{Number of URLs Used} (evidence incorporation), \emph{Character Count} (note length, assessing platform compatibility), and \emph{Number of Notes} generated (pipeline success out of 488 total).
\rr{We assess the quality of cited sources using Media Bias/Fact Check (MBFC)\footnote{\url{https://mediabiasfactcheck.com/}} ratings. For each cited domain, we assign a binary Domain Credibility score and map MBFC's political bias labels (Far Left to Far Right) to a Domain Bias score ranging from $-3$ to $+3$, with Least Biased mapped to $0$. We then average both scores across the cited domains in each note, excluding domains not listed in MBFC.}

\paragraph{LLM-as-a-Judge:}
We use GPT-5.2 as a judge to assess notes across five dimensions on a Likert scale of 1--5 where higher is better: \emph{Functional Errors}; \emph{Claim Alignment}; \emph{Fact Alignment}; \emph{Completeness}; and \emph{Helpfulness} (Appendix \ref{app:prompt_temp}). \rr{The judge receives the original tweet, a fixed human-written \texttt{CRH} reference note, the candidate note, and a dimension-specific rubric, without web retrieval. It evaluates consistency and alignment with the supplied information rather than independently discovering or verifying claims. This follows established formulations of factual consistency as text-to-text information alignment \cite{zha2023alignscore}; providing reference answers and explicit evaluation criteria has also been shown to improve evaluator–human correspondence \cite{kim2024prometheus}. Furthermore, we discuss the alignment between LLM vs. Human judges in Appendix \ref{app:llmaaj_human_align}.}

\paragraph{Human Evaluation:}
To validate automated assessments, we conduct human evaluation on a representative sample of AI-generated notes against potentially misleading tweets. Following the Community Notes framework, annotators assessed three core dimensions of usefulness: \emph{Factuality} (verifying correctness, source accessibility, direct claim support from sources, and source credibility), \emph{Completeness} (evaluating relevancy to the tweet's misleading content and inclusion of all necessary context), and \emph{Helpfulness} (holistic 1--5 rating of overall usefulness considering clarity, neutrality, relevant context, and reliable sourcing). The annotation guidelines are provided in Appendix \ref{app:he_guide}. 

\section{The \ds\ Dataset}
\label{sec:dataset}
We construct \ds{}, a dataset of English-language tweets concerning U.S. politics accompanied by their corresponding Community Notes. The data collection is based on a two-stage pipeline that begins with Community Note identification and subsequently retrieves the tweets they reference. This design ensures that every tweet in the dataset is grounded in observed, user-generated fact-checking activity rather than inferred relevance or synthetic labels.

Our dataset integrates two primary data sources. The first is X's Community Notes, dataset\footnote{\url{https://x.com/i/communitynotes/download-data}} which encompasses notes, ratings, note status history, and user enrollment status, capturing contributions through September 7, 2025. Second, we used the X API to obtain the tweet text corresponding to the Community Notes data.

In order to isolate U.S. political content, we implemented \textbf{topical filtering} at the Community Note level before large-scale tweet retrieval. We adopted a two-step, high-precision filtering procedure: \textbf{(1)~Zero-shot classification} using an MNLI-based model\footnote{\url{https://hf.co/FacebookAI/roberta-large-mnli}} to identify notes related to U.S. politics, and \textbf{(2)~Keyword-based tagging} applying a curated set of U.S. political keywords\footnote{\url{https://www.factcheck.org/archives/}} to capture relevant notes overlooked by the classifier. We retain the union of both filters to enhance recall while maintaining precision. All tweets linked to at least one filtered note are retained. Quality validation on a manually annotated subset demonstrates that this hybrid strategy achieves substantially higher recall for U.S. political notes than either method alone (refer to Appendix \ref{app:ds_filteration}). \textbf{(3) Additional filtering} ensured textual comparability and contextual completeness. First, we excluded tweets whose associated notes indicated that essential context depends on images or videos. Second, we removed tweets written in non-English languages, even when notes are in English. Third, we excluded tweets posted before 2021, as Community Notes' coverage before this period is extremely sparse.

After filtering, the dataset contains \emph{78,698} unique English-language tweets with associated Community Notes, which form the basis for all analysis and evaluation. The dataset is highly imbalanced: $\sim$92\% of notes remain in the \texttt{NMR} state, with only 3.57\% rated helpful and 4.19\% rated unhelpful. Notes are short (28 words on average), in many cases lack citations (22.93\%), and exhibit substantial temporal delays, with a 5h median lag between tweet and note creation and resolution. Detailed statistics are provided in Appendix \ref{app:ds_stats}.

For evaluation, we use temporally stratified sampling, selecting the most recent 10\% of tweets with at least one \texttt{CRH} note. This yields 488 tweets posted between May 1 and September 6, 2025, which serve as reference points for evaluation. Only 265 of these contain at least one \texttt{NMR} note, highlighting a key limitation of Supernotes in addressing emerging misinformation.

\section{\rr{Results}}
\label{sec:res_dis}
We investigate whether gap-based retrieval can expand coverage without compromising note quality. We first compare \ours{} with note-dependent and unconditioned web-search baselines, then validate the findings through human evaluation and ablation studies. As auxiliary checks, we also examine traditional similarity-based evaluation measures. We observe overall lower scores, reflecting paraphrasing and independent evidence sourcing. Additionally, all systems cite credible domains and show Left--Center political leaning, which is often observed in the literature as well \cite{kuuse2025crowdsourced}. Full auxiliary results are provided in Appendix~\ref{app:auto_eval}.

\subsection{Gap-based Retrieval Improves Quality}
\paragraph{Resolving the coverage--quality trade-off.}
Table~\ref{tab:auto_evals_main} (and Appendix~\ref{app:auto_eval}) reveals a central tension between note-dependent and web-search-based generation. Supernotes-Lite achieves strong LLM judge-based scores but can operate only on the 265 tweets containing existing \texttt{NMR} notes, covering 54\% of the test set. Web-Agents remove this dependency and achieve near-complete coverage, but their notes are consistently less aligned with the claims and contextual requirements of the corresponding human-written notes. In contrast, \ours{} generates notes for 99--100\% of the test set while recovering much of the quality associated with note-dependent generation. For example, with Gemini-2.5-Flash, \ours{} achieves a \textit{Claim Alignment} score of 3.744 at 99\%, compared with 3.898 for Supernotes-Lite at 54\% and 3.315 for the Web-Agent at near-complete coverage.

\begin{table}[t]
\centering
\resizebox{\columnwidth}{!}{
\renewcommand{\arraystretch}{1.2}
\begin{tabular}{c|lccccc}
\cline{1-2} \cline{3-7} 
 &  & \multicolumn{5}{c}{\textbf{LLM-as-a-Judge}}  \\ \cline{3-7}
\multirow{-2}{*}{\rotatebox[origin=c]{90}{\textbf{Type}}} & \multirow{-2}{*}{\textbf{Model}} & \textbf{FE $\bm \uparrow$} & \textbf{CA $\bm \uparrow$} & \textbf{FA $\bm \uparrow$} & \textbf{C $\bm \uparrow$} & \textbf{H $\bm \uparrow$} \\ \cline{1-2} \cline{3-7} 
 & Llama-3.1-8B & 4.087 & 3.585 & 2.479 & 2.464 & 2.796 \\
 & Ministral-8B & 4.083 & 3.589 & 2.487 & 2.355 & 2.672 \\
 & Qwen3-14B & 3.974 & 3.728 & 2.615 & 2.487 & 2.838 \\
 & Apriel-Nemotron-15B & 3.879 & 3.709 & 2.540 & 2.528 & 2.811 \\
 & GPT-5-nano & 4.053 & 3.668 & 2.634 & 2.506 & 2.860 \\
\multirow{-6}{*}{\rotatebox[origin=c]{90}{Supernote Lite (54\%)}} & Gemini-2.5-Flash & 4.358 & \textbf{3.898} & \textbf{2.906} & 2.687 & 3.128 \\ \cline{1-2} \cline{3-7} 
 & GPT-5-nano (T1) & \textbf{4.516} & 3.383 & 2.488 & 2.459 & 2.961 \\
 & GPT-5-nano (T2) & 4.496 & 3.340 & 2.465 & 2.426 & 2.984 \\
 & \cellcolor[HTML]{EFEFEF}$^{\text{GPT-5-nano}}\Delta_{\text{WA}_{T2}-\text{WA}_{T1}}$ & \cellcolor[HTML]{EFEFEF}-0.02 & \cellcolor[HTML]{EFEFEF}-0.043 & \cellcolor[HTML]{EFEFEF}-0.023 & \cellcolor[HTML]{EFEFEF}-0.033 & \cellcolor[HTML]{EFEFEF}0.023 \\
 & \cellcolor[HTML]{EFEFEF}$^{\text{GPT-5-nano}}\Delta_{\text{WA}_{T2}-\text{SN}}$ & \cellcolor[HTML]{EFEFEF}0.443\tiny{\textsuperscript{***}} & \cellcolor[HTML]{EFEFEF}-0.328\tiny{\textsuperscript{***}} & \cellcolor[HTML]{EFEFEF}-0.169 & \cellcolor[HTML]{EFEFEF}-0.08 & \cellcolor[HTML]{EFEFEF}0.124 \\
 & Gemini-2.5-Flash (T1) & 4.242 & 3.334 & 2.465 & 2.418 & 2.809 \\
 & Gemini-2.5-Flash (T2) & 4.127 & 3.315 & 2.465 & 2.388 & 2.772 \\
 & \cellcolor[HTML]{EFEFEF}$^{\text{Gemini-2.5-Flash}}\Delta_{\text{WA}_{T2}-\text{WA}_{T1}}$ & \cellcolor[HTML]{EFEFEF}-0.115 & \cellcolor[HTML]{EFEFEF}-0.019 & \cellcolor[HTML]{EFEFEF}0 & \cellcolor[HTML]{EFEFEF}-0.03 & \cellcolor[HTML]{EFEFEF}-0.037 \\
 & \cellcolor[HTML]{EFEFEF}$^{\text{Gemini-2.5-Flash}}\Delta_{\text{WA}_{T2}-\text{SN}}$ & \cellcolor[HTML]{EFEFEF}-0.231\tiny{\textsuperscript{***}} & \cellcolor[HTML]{EFEFEF}-0.583\tiny{\textsuperscript{***}} & \cellcolor[HTML]{EFEFEF}-0.441\tiny{\textsuperscript{***}} & \cellcolor[HTML]{EFEFEF}-0.299\tiny{\textsuperscript{**}} & \cellcolor[HTML]{EFEFEF}-0.356\tiny{\textsuperscript{***}} \\
 & Grok-4 (T1) & 1.568 & 3.213 & 2.342 & 2.236 & 1.795 \\
\multirow{-10}{*}{\rotatebox[origin=c]{90}{Web Agent (100\%)}} & Sonar-Deep-Res (T1) & 1.816 & 3.318 & 2.393 & 2.416 & 1.994 \\ \cline{1-2} \cline{3-7} 
 & GPT-5-nano (T2) & 4.266 & 3.668 & 2.656 & 2.734 & \textbf{3.141} \\
 & \cellcolor[HTML]{EFEFEF}$^{\text{GPT-5-nano}}\Delta_{\text{OUR}_{T2}-\text{WA}_{T2}}$ & \cellcolor[HTML]{EFEFEF}{\color[HTML]{FE0000} -0.23\tiny{\textsuperscript{***}}} & \cellcolor[HTML]{EFEFEF}{\color[HTML]{009901} 0.328\tiny{\textsuperscript{***}}} & \cellcolor[HTML]{EFEFEF}{\color[HTML]{009901} 0.191\tiny{\textsuperscript{*}}} & \cellcolor[HTML]{EFEFEF}{\color[HTML]{009901} 0.308\tiny{\textsuperscript{***}}} & \cellcolor[HTML]{EFEFEF}{\color[HTML]{009901} 0.157} \\
 & \cellcolor[HTML]{EFEFEF}$^{\text{GPT-5-nano}}\Delta_{\text{OUR}_{T2}-\text{SN}}$ & \cellcolor[HTML]{EFEFEF}{\color[HTML]{009901} 0.213\tiny{\textsuperscript{***}}} & \cellcolor[HTML]{EFEFEF}0 & \cellcolor[HTML]{EFEFEF}{\color[HTML]{009901} 0.022} & \cellcolor[HTML]{EFEFEF}{\color[HTML]{009901} 0.228\tiny{\textsuperscript{*}}} & \cellcolor[HTML]{EFEFEF}{\color[HTML]{009901} 0.281\tiny{\textsuperscript{**}}} \\
 & Gemini-2.5-Flash (T2) & 4.477 & 3.744 & 2.682 & \textbf{2.744} & 3.114 \\
 & \cellcolor[HTML]{EFEFEF}$^{\text{Gemini-2.5-Flash}}\Delta_{\text{OUR}_{T2}-\text{WA}_{T2}}$ & \cellcolor[HTML]{EFEFEF}{\color[HTML]{009901} 0.35\tiny{\textsuperscript{***}}} & \cellcolor[HTML]{EFEFEF}{\color[HTML]{009901} 0.429\tiny{\textsuperscript{***}}} & \cellcolor[HTML]{EFEFEF}{\color[HTML]{009901} 0.217\tiny{\textsuperscript{**}}} & \cellcolor[HTML]{EFEFEF}{\color[HTML]{009901} 0.356\tiny{\textsuperscript{***}}} & \cellcolor[HTML]{EFEFEF}{\color[HTML]{009901} 0.342\tiny{\textsuperscript{***}}} \\
\multirow{-6}{*}{\rotatebox[origin=l]{90}{\ours\ (99\%)}} & \cellcolor[HTML]{EFEFEF}$^{\text{Gemini-2.5-Flash}}\Delta_{\text{OUR}_{T2}-\text{SN}}$ & \cellcolor[HTML]{EFEFEF}{\color[HTML]{009901} 0.119\tiny{\textsuperscript{*}}} & \cellcolor[HTML]{EFEFEF}{\color[HTML]{FE0000} -0.154} & \cellcolor[HTML]{EFEFEF}{\color[HTML]{FE0000} -0.224\tiny{\textsuperscript{*}}} & \cellcolor[HTML]{EFEFEF}{\color[HTML]{009901} 0.057} & \cellcolor[HTML]{EFEFEF}{\color[HTML]{FE0000} -0.014} \\ \cline{1-2} \cline{3-7} 
\end{tabular}
}
\caption{\rr{LLM-as-a-Judge results for AI-generated CNs, using human-written helpful notes as ground truth. Here, \textbf{Type} represents generation method (coverage/success rate) and  \textbf{Scores (1--5):} Functional Errors (FE), Claim Alignment (CA), Fact Alignment (FA), Completeness (C), and Helpfulness (H). Gray-shaded rows report differences ($\Delta$); {\color[HTML]{009901}\textbf{green}}/{\color[HTML]{FE0000}\textbf{red}} coloring is shown only for \ours{} comparisons. Statistical significance (using independent t-test), where *,**,***, $p<\{0.05,0.01,0.001\}$, respectively.}}
\label{tab:auto_evals_main}
\vspace{-\baselineskip}
\end{table}

\paragraph{Gap conditioning improves unconditioned web search.}
The comparison between \ours{} and Web-Agent directly isolates the effect of gap-conditioned retrieval. Both systems use the same backbone, web-search access, and single-call inference budget, differing only in whether retrieval is guided by the identified information gaps. With Gemini-2.5-Flash, gap conditioning significantly improves \textit{Claim Alignment} (+0.429), \textit{Fact Alignment} (+0.217), \textit{Completeness} (+0.356), and \textit{Helpfulness} (+0.342; all $p<0.01$). GPT-5-nano exhibits the same pattern, with significant gains in \textit{Claim Alignment} (+0.328), \textit{Fact Alignment} (+0.191), and \textit{Completeness} (+0.308). These controlled comparisons show that the quality deficit of generic Web-Agents arises from how retrieval is directed rather than from insufficient model capability or information access.

\paragraph{Quality relative to note-dependent generation.}
Despite operating over a substantially broader set of tweets, \ours{} remains competitive with Supernotes-Lite. With GPT-5-nano, \ours{} significantly improves \textit{Completeness} (+0.228, $p<0.05$) and \textit{Helpfulness} (+0.281, $p<0.01$). With Gemini-2.5-Flash, the differences in \textit{Claim Alignment}, \textit{Completeness}, and \textit{Helpfulness} are not statistically significant, although \textit{Fact Alignment} is lower by 0.224 ($p<0.05$). Thus, the two approaches achieve largely comparable aggregate quality in their respective operating regimes, while \ours{} extends generation to nearly all of the 46\% of tweets that Supernotes-Lite cannot address.

\begin{figure*}[!t]
    \centering
    \begingroup
    \renewenvironment{figure}[1][]{\par\centering}{\par}

    \begin{minipage}[t]{0.33\textwidth}
        \centering
        \input{figures/he_win_rate}
    \end{minipage}
    \hfill
    \begin{minipage}[t]{0.315\textwidth}
        \centering
        \input{figures/gt_vs_scores}
    \end{minipage}
    \hfill
    \begin{minipage}[t]{0.28\textwidth}
        \centering
        \input{figures/temporal_scores}
    \end{minipage}
    \endgroup
\end{figure*}

\subsection{Human Evaluation Favors \ours{}}
Furthermore, we perform human evaluation for fine-grained reference-independent analysis. Two evaluators with working knowledge of U.S.\ politics and research experience in CSS/NLP each assessed notes for 50 randomly sampled tweets, yielding 100 tweets and 400 evaluated notes across four conditions. All AI systems use Gemini-2.5-Flash. Table~\ref{tab:human_eval_res} reports dimension-level scores, while Figure~\ref{fig:he_win_rate} presents pairwise preferences. 

\begin{table}[!t]
\centering
\resizebox{\columnwidth}{!}{
\renewcommand{\arraystretch}{1.3}
\begin{tabular}{lccccccc}
\hline
\textbf{Method} & \textbf{F1 $\bm \uparrow$} & \textbf{F2 $\bm \uparrow$} & \textbf{F3 $\bm \uparrow$} & \textbf{F4 $\bm \uparrow$} & \textbf{C1 $\bm \uparrow$} & \textbf{C2 $\bm \uparrow$} & \textbf{H $\bm \uparrow$} \\ \hline
Human & 0.94 & 0.91 & 0.87 & 0.81 & 0.93 & 0.69 & 3.36 \\
Supernote Lite & 0.96 & 0.83 & 0.73 & 0.77 & 0.91 & 0.53 & 3.14 \\
Web Agent & 0.96 & 0.80 & 0.89 & 0.88 & 0.85 & 0.65 & 3.44 \\
\ours{} & 0.99 & 0.89 & 0.91 & 0.90 & 0.97 & 0.87 & 3.87 \\
\rowcolor[HTML]{EFEFEF} 
$\Delta_{\text{OUR}-\text{Human}}$ & {\color[HTML]{009901} 0.05} & {\color[HTML]{FE0000} -0.02} & {\color[HTML]{009901} 0.04} & {\color[HTML]{009901} 0.09} & {\color[HTML]{009901} 0.04} & {\color[HTML]{009901} 0.18**} & {\color[HTML]{009901} 0.51**} \\ \hline
\end{tabular}}

\caption{Human evaluation of AI-generated Community Notes. Factuality: F1 (factual accuracy), F2 (source accessibility), F3 (source support), F4 (source credibility). Completeness: C1 (relevance), C2 (context coverage). H: Helpfulness (1–5 scale).}
\label{tab:human_eval_res}
\vspace{-\baselineskip}
\end{table}

\paragraph{Human prefer \ours{} to \texttt{CRH} notes.}
\ours{} is preferred to human-written helpful notes in 69\% of pairwise comparisons and receives significantly higher overall helpfulness scores (3.87 vs.\ 3.36, $p<0.01$). The advantage is primarily attributable to more comprehensive context coverage: \ours{} scores 0.87 on C2, compared with 0.69 for human-written notes ($p<0.01$). Factual accuracy remains comparable across the two conditions (F1: 0.99 vs.\ 0.94), indicating that the improvement arises from supplying more of the context needed to interpret the misleading claim rather than from a difference in basic factual correctness. Human writers might be constrained by the efforts required to provide comprehensive context even when the central correction is accurate.

Given that \ours{} notes are generally longer, we test whether this helpfulness advantage can be explained by verbosity. The \ours{} effect remains positive and significant after controlling for note length, whereas length itself does not significantly predict either helpfulness or pairwise preference (Appendix~\ref{app:length_robustness}).

\paragraph{Retrieval structure improves upon unconditioned access.}
When compared with the Web-Agent using the same model and web-search budget, \ours{} achieves a 59\% pairwise win rate and a 0.43-point improvement in helpfulness (3.87 vs.\ 3.44). Since model capability, information access, and inference budget are held fixed, this comparison isolates the benefit of organizing retrieval around explicit information gaps. The result corroborates the automatic evaluation that access to web evidence alone is insufficient unless retrieval is directed toward the contextual and evidentiary deficits that determine note usefulness.

\paragraph{Human judgments capture gains missed by reference metrics.}
Evaluators prefer \ours{} to Supernotes-Lite in 81\% of pairwise comparisons, despite the closer automatic scores of the two systems. This divergence indicates that reference-based evaluation underestimates notes that fulfill the same communicative objective through different wording and independently retrieved evidence. Human evaluation instead rewards the additional context and direct claim coverage introduced by gap-conditioned retrieval.

\subsection{Ablation Study}
\label{sec:abl_study}
\paragraph{Bridging cold-start coverage and warm-start quality.}
Table~\ref{tab:gitsearch_abl} evaluates \ours{} in the two regimes motivating our framework. When $\mathcal{C}_i=\emptyset$, Supernotes-Lite cannot operate, while Web-Agents preserve coverage but sacrifice quality; in this cold-start setting, \ours{} generally improves over the matched Web-Agent, showing that gap-guided retrieval remains effective without prior crowd input. When \texttt{NMR} notes are available, \ours{} is competitive with Supernotes-Lite and can further exploit these incomplete notes as signals of what information is still missing. For Gemini-2.5-Flash, adding $\mathcal{C}_i$ significantly improves \textit{Claim Alignment} (+0.322), \textit{Fact Alignment} (+0.275), \textit{Completeness} (+0.255), and \textit{Helpfulness} (+0.225; all $p<0.05$). These results show that \ours{} preserves generation in cold-start where note-dependent methods cannot run, while using partial crowd context when available to improve note quality.

\begin{table}[!t]
\resizebox{\columnwidth}{!}{
\begin{tabular}{llccccc}
\hline
 & \textbf{Method} & \textbf{FE $\bm \uparrow$} & \textbf{CA $\bm \uparrow$} & \textbf{FA $\bm \uparrow$} & \textbf{C $\bm \uparrow$} & \textbf{H $\bm \uparrow$} \\ \hline
 & Web Agent ($\mathcal{C}_i = \emptyset$) & 4.475 & 3.345 & 2.457 & 2.435 & 2.991 \\
 & \ours{} ($\mathcal{C}_i = \emptyset$) & 4.238 & 3.578 & 2.565 & 2.623 & 3.049 \\
 & \cellcolor[HTML]{EFEFEF}$^{\mathcal{C}_i = \emptyset}\Delta_{OUR-WA}$ & \cellcolor[HTML]{EFEFEF}{\color[HTML]{FE0000} -0.237\tiny{\textsuperscript{***}}} & \cellcolor[HTML]{EFEFEF}{\color[HTML]{009901} 0.233} & \cellcolor[HTML]{EFEFEF}{\color[HTML]{009901} 0.108} & \cellcolor[HTML]{EFEFEF}{\color[HTML]{009901} 0.188} & \cellcolor[HTML]{EFEFEF}{\color[HTML]{009901} 0.058} \\
 & Supernote Lite ($\mathcal{C}_i \neq \emptyset$) & 4.053 & 3.668 & 2.634 & 2.506 & 2.86 \\
 & \ours{} ($\mathcal{C}_i \neq \emptyset$) & 4.291 & 3.743 & 2.732 & 2.826 & 3.219 \\
 & \cellcolor[HTML]{EFEFEF}$^{\mathcal{C}_i \neq \emptyset}\Delta_{OUR-SN}$ & \cellcolor[HTML]{EFEFEF}{\color[HTML]{009901} 0.238\tiny{\textsuperscript{***}}} & \cellcolor[HTML]{EFEFEF}{\color[HTML]{009901} 0.075} & \cellcolor[HTML]{EFEFEF}{\color[HTML]{009901} 0.098} & \cellcolor[HTML]{EFEFEF}{\color[HTML]{009901} 0.32\tiny{\textsuperscript{**}}} & \cellcolor[HTML]{EFEFEF}{\color[HTML]{009901} 0.359\tiny{\textsuperscript{***}}} \\ \vspace{-3.5mm}
 & & & & & & \\ 
\multirow{-8}{*}{\rotatebox[origin=c]{90}{GPT-5-nano}} & \cellcolor[HTML]{EFEFEF}$^{OUR}\Delta_{(\mathcal{C}_i \neq \emptyset) - (\mathcal{C}_i = \emptyset)}$ & \cellcolor[HTML]{EFEFEF}{\color[HTML]{009901} 0.053} & \cellcolor[HTML]{EFEFEF}{\color[HTML]{009901} 0.165} & \cellcolor[HTML]{EFEFEF}{\color[HTML]{009901} 0.167} & \cellcolor[HTML]{EFEFEF}{\color[HTML]{009901} 0.203} & \cellcolor[HTML]{EFEFEF}{\color[HTML]{009901} 0.17} \\ \hline
 & Web Agent ($\mathcal{C}_i = \emptyset$) & 4.186 & 3.364 & 2.545 & 2.436 & 2.827 \\
 & \ours{} ($\mathcal{C}_i = \emptyset$) & 4.486 & 3.568 & 2.532 & 2.605 & 2.991 \\
 & \cellcolor[HTML]{EFEFEF}$^{\mathcal{C}_i = \emptyset}\Delta_{OUR-WA}$ & \cellcolor[HTML]{EFEFEF}{\color[HTML]{009901} 0.3\tiny{\textsuperscript{***}}} & \cellcolor[HTML]{EFEFEF}{\color[HTML]{009901} 0.204} & \cellcolor[HTML]{EFEFEF}{\color[HTML]{FE0000} -0.013} & \cellcolor[HTML]{EFEFEF}{\color[HTML]{009901} 0.169} & \cellcolor[HTML]{EFEFEF}{\color[HTML]{009901} 0.164} \\
 & Supernote Lite ($\mathcal{C}_i \neq \emptyset$) & 4.358 & 3.898 & 2.906 & 2.687 & 3.128 \\
 & \ours{} ($\mathcal{C}_i \neq \emptyset$) & 4.47 & 3.89 & 2.807 & 2.86 & 3.216 \\
 & \cellcolor[HTML]{EFEFEF}$^{\mathcal{C}_i \neq \emptyset}\Delta_{OUR-SN}$ & \cellcolor[HTML]{EFEFEF}{\color[HTML]{009901} 0.112} & \cellcolor[HTML]{EFEFEF}{\color[HTML]{FE0000} -0.008} & \cellcolor[HTML]{EFEFEF}{\color[HTML]{FE0000} -0.099} & \cellcolor[HTML]{EFEFEF}{\color[HTML]{009901} 0.173} & \cellcolor[HTML]{EFEFEF}{\color[HTML]{009901} 0.088} \\ \vspace{-3.5mm}
 & & & & & & \\ 
\multirow{-8}{*}{\rotatebox[origin=c]{90}{Gemini-2.5-Flash}} & \cellcolor[HTML]{EFEFEF}$^{OUR}\Delta_{(\mathcal{C}_i \neq \emptyset) - (\mathcal{C}_i = \emptyset)}$ & \cellcolor[HTML]{EFEFEF}{\color[HTML]{FE0000} -0.016} & \cellcolor[HTML]{EFEFEF}{\color[HTML]{009901} 0.322\tiny{\textsuperscript{**}}} & \cellcolor[HTML]{EFEFEF}{\color[HTML]{009901} 0.275\tiny{\textsuperscript{*}}} & \cellcolor[HTML]{EFEFEF}{\color[HTML]{009901} 0.255\tiny{\textsuperscript{*}}} & \cellcolor[HTML]{EFEFEF}{\color[HTML]{009901} 0.225\tiny{\textsuperscript{*}}} \\ \hline
\end{tabular}}
\caption{\rr{Ablation study quantifying the impact of including existing community notes ($\mathcal{C}_i$) as context.}}
\label{tab:gitsearch_abl}
\vspace{-\baselineskip}
\end{table}


\paragraph{Robustness across inference time.}
Since web evidence and search rankings change over time, we test whether the weaker Web-Agent results reflect a disadvantage associated with its original inference timestamp. We rerun the Web-Agent baselines $\sim$3 months later and compare their outputs at $T_1$ and $T_2$. Neither GPT-5-nano nor Gemini-2.5-Flash exhibits a significant change on any LLM-as-a-Judge dimension. The performance gap between \ours{} and the corresponding Web-Agent therefore persists across inference timestamps and cannot be explained by transient differences in search recency. This temporal replication strengthens the conclusion that the observed improvement is attributable to gap-conditioned retrieval rather than to a favorable retrieval snapshot.

\section{Diagnostic Analysis}
\label{sec:diag_analysis}
For fine-grained diagnostic analysis, we analyze variation across information-gap types, sensitivity to the maturity of available web evidence, and representative failure cases. Here, we use outputs from Gemini-2.5-Flash backbone.

\subsection{Performance Varies by Gap Type}
Figure \ref{fig:gt_vs_scores} disaggregates LLM-as-a-Judge evaluation by gap type, revealing systematic differences in verification difficulty. Gaps involving contradictions and unsubstantiated claims perform the best across evaluation measures, with \textit{Claim Alignment} (CA) scores above 3.8 and \textit{Helpfulness} (H) around 3.2–3.5, indicating that explicitly stated and locally verifiable claims are most tractable for the model. In contrast, missing context and missing coverage gaps show substantially lower scores, particularly on \textit{Factual Alignment} and \textit{Completeness} (FA/C $\approx$ 2.2–2.3), reflecting difficulty in synthesizing dispersed or implicit evidence. Performance is lowest for vague reference across all criteria (e.g., CA $\approx$ 3.27, FA $\approx$ 1.91, C $\approx$ 2.27), highlighting ambiguity resolution as a persistent bottleneck.

\subsection{Sensitivity to Evidence Maturity}
Human-written notes are contemporaneous with their tweets, whereas \ours{} retrieves evidence post-hoc and may benefit from hindsight. We probe this effect by disaggregating LLM-as-a-Judge scores by tweet month (Figure~\ref{fig:temporal_scores}). Helpfulness is highest for May tweets ($H=3.51$) and lower for June--August ($H=2.98$--$3.16$)\footnote{Gemini-2.5-Flash's January 2025 training cutoff precedes our May--September evaluation window.}. Although the trend is not monotonic, suggesting interactions with topic and gap-type composition, it is consistent with evidence consolidating over time~\cite{vu2024freshllms}. We discuss the implications for real-time deployment in the Limitations section.

\subsection{Case Study}
Representative cases from human evaluation (Appendix~\ref{app:case_study_examples}) reveal three recurring error patterns. First, the baseline systems occasionally produce notes that are factually correct but \textit{contextually misaligned}, e.g., citing general legal prohibitions rather than engaging with the specific legislative claim in the tweet. Second, for underspecified posts (e.g., ``FIRED.''), systems identify missing context but stop short of resolving the claim, yielding low-value fact-checks. Third, systems that defer to tool outputs for exhaustive claims (e.g., ``no record exists'') produce weaker corrections, even when contradictory evidence is available. \ours{}-specific errors are narrower. It occasionally provides accurate but less exhaustive counterevidence than human or Web-Agent notes, and struggles with vague references requiring conversational or temporal reconstruction (Figure~\ref{fig:gt_vs_scores}; FA $\approx$ 1.9). Intermediate pipeline outputs illustrating these patterns are in Appendix~\ref{app:intermediate_outputs}.

\section{Conclusion}
\label{sec:conc}
\rr{We have presented \ours{}, a framework that identifies information gaps, uses them to guide web search, and generates evidence-grounded Community Notes. We have also introduced \ds{}, a dataset of 78,698 U.S. political tweets and their associated Community Notes. Our experiments show that \ours{} achieves 99\% coverage, compared with 54\% for Supernotes-Lite, while maintaining comparable note quality. In human evaluation, annotators preferred \ours{} over general web-search agents in 59\% of comparisons and over human-written helpful notes in 69\%. We further found that even incomplete Community Notes can help guide retrieval, although vague references remain challenging. In future work, we plan to evaluate the framework at different stages of a post's lifecycle and explore adaptive gap detection as new evidence becomes available.}

\section*{Limitations}
\ours{} is evaluated only on English-language U.S. political tweets; generalizability to other languages, domains (e.g., health misinformation), or political contexts remains untested, and our gap taxonomy (Section~\ref{sec:framework}), though empirically validated here, may need re-derivation elsewhere. \rr{A manual review of randomly sampled cases also found that the taxonomy struggles when claims are backed by abundant but low-quality or misleading sources: the deficit is one of evidence \textit{credibility} rather than a missing gap category. Incorporating credibility-aware signals, e.g.,~by scaling resources like Media Bias/Fact Check, is a natural extension.}

Phase~II's single-call retrieval, while efficient, can under-resolve complex multi-hop claims and inherits source bias toward crawlable outlets~\cite{fletcher2024how}; our error analysis (Figure~\ref{fig:gt_vs_scores}) shows \textit{vague reference} and \textit{missing context} gaps remain hardest, as they require discourse-level reasoning beyond a single retrieval pass. \rr{We also find lower performance on the most recent events in our test window, suggesting that the maturity of a topic's online evidence base -- not retrieval access alone -- affects note quality, and that \ours{} may be less reliable for breaking claims than our aggregate metrics suggest.}

Phase~III's grounding constraint is enforced via prompting rather than architecture, so compliance may degrade on adversarial or out-of-distribution inputs; empirical URL recall (Table~\ref{tab:auto_evals}) and human evaluation (Table~\ref{tab:human_eval_res}) suggest it holds in practice, but residual risks of factual error and framing bias remain for ambiguous gaps, and the pipeline adds inference latency over single-call baselines. These factors support deploying \ours{} with human oversight rather than as a fully autonomous moderator; future work could explore iterative, adaptive retrieval that refines gaps as evidence accumulates~\cite{li2024re}, using partial crowd contributions as structured inputs to gap detection.

\section*{Ethical Statement}
We adhere to the ethical guidelines regarding the use of social media data and AI-generated content. All data used in this study, including the \ds{} dataset, was obtained from publicly available sources via the APIs and the X's Community Notes data archive. We acknowledge that the dataset contains real-world political discourse, which may include toxic or offensive content; these examples are retained solely for the purpose of evaluating moderation systems. Additionally, this study and its human evaluation protocol received an exemption from the Ethics Review Committee of our University, given the use of publicly available data and minimal risk to participants. \rr{Consistent with X's developer terms, we redistribute note and tweet identifiers with hydration scripts rather than tweet text (Appendix~\ref{app:code}).}

We acknowledge that the pipeline could theoretically be repurposed to generate plausible-sounding misinformation. We mitigate this by (a) grounding all outputs in retrieved evidence with verifiable URLs, and (b) recommending human-in-the-loop deployment. Furthermore, while \ours{} framework significantly reduces hallucinations through evidence-based grounding, but AI-generated fact-checks can still carry inherent risks of bias or inaccuracy. Consequently, we advocate for the deployment of this framework as a ``human-in-the-loop'' assistive tool to augment human moderators, rather than as a fully autonomous decision-making system.

\section*{Acknowledgments}
This research is supported by the Ministry of Education (MOE), Singapore, through its AcRF TIER 3 Grant (MOE-MOET32022-0001). We greatly acknowledge the time and effort of our human evaluators. We thank Svetlana Churina, Insyirah Mujtahid, Riddhanya Senapathi, and Jiaying Wu for early conversations related to the Community Notes concepts.
\paragraph{AI disclaimer:} We acknowledge the use of an AI-based tool, Grammarly, solely for polishing and improving the clarity of our writing. All analysis, framing, coding, and other substantive aspects of the work were conducted independently by the authors.

\bibliography{custom}

\appendix
\rr{
\section{Gap Taxonomy Details}
\label{app:taxo_details}
\subsection{Definition of Gap Types}
\label{app:def_gaps}
We define six gap types identified during Phase I (Gap Detection \& Prioritization). Each gap type targets a distinct pattern of verification failure observed in Community Notes.
\paragraph{Unsubstantiated Claim.} A factual assertion made in the tweet without any accompanying evidence or citation. Example: a claim like ``\emph{studies show X}'' with no study referenced. When no existing notes are available (cold start), every factual assertion in the tweet is treated as an unsubstantiated claim by default.
\paragraph{Contradiction.} Conflicting information either across candidate notes (e.g., one note claims an effect ``\emph{increases}'' while another claims it ``\emph{decreases}'') or within the tweet itself (e.g., internal inconsistency or a logical fallacy), when no notes exist.
\paragraph{Vague Reference.} A non-specific mention of evidence, legislation, events, or entities that must be resolved to a concrete referent before verification can proceed. Example: ``\emph{some studies}'' or ``\emph{recent reports}'' without further detail.
\paragraph{Missing Context.} A statistic, number, or claim presented without the surrounding information (e.g., baseline, timeframe, or location) needed to interpret it correctly. Example: ``\emph{crime increased 50\%}'' without stating over what period or where.
\paragraph{Source Verification.} A source is named but not linked, or a cited source requires independent confirmation that it exists and supports the claim. Example: ``\emph{according to a Harvard study}'' with no accessible citation.
\paragraph{Missing Coverage.} An aspect of the tweet's claim(s) that existing notes fail to address. Example: the tweet makes three distinct claims, but candidate notes address only one. In cold-start settings, this captures cases where the tweet's broader narrative lacks any contextualization or fact-check.
~
\subsection{Completeness of Gap Taxonomy} 
\label{app:gap_completeness}
We randomly sample 100 tweet-notes pairs to assess the completeness of the proposed taxonomy. We find the taxonomy is sufficient to represent the information deficits needed for targeted search in 82\% of cases. Among the remaining 18\%, the dominant limitation concerned evidence quality and credibility, where the deficit arose not from missing categories but from claims supported by low-quality or misleading evidence (see Limitations).
~
\subsection{Accuracy of Gap Identification}
\label{app:gap_accuracy}
To assess the accuracy and reliability of LLMs in extracting gaps from tweet–note pairs, we randomly sample 100 instances generated by Gemini-2.5-Flash. We find that the identified gaps are accurate and well-grounded in 85.4\% of the cases. Furthermore, to assess the reliability of the accuracy annotations, we calculate Krippendorff’s Alpha and obtain a score of 0.79, indicating substantial agreement between our two independent annotators \cite{marzi2024kalpha}.
}

\section{Experimental Details}
\label{app:exp_details}
\subsection{Dataset \& Code Availability}
\label{app:code}
We release note IDs, tweet IDs, all Community Notes metadata, and hydration scripts; tweet text is not redistributed, in compliance with X's developer terms. At collection time, 12.1\% of tweets were already deleted; released IDs reflect this.

\subsection{\ds{}'s Filteration Quality Assessment}
\label{app:ds_filteration}
We begin with a quality check on a 1K sample of filtered notes using a two-class classifier (US politics vs Others). A human annotator with expertise in US politics labels the data, and our hybrid method reaches a macro F1 of 0.892. Its main gain is the much higher recall for US-politics notes, rising to 0.710 compared with 0.428 for MNLI and 0.524 for keywords. As shown in Table \ref{tab:filtered_data_quality}, this reduces the number of missed political notes and gives us a cleaner dataset overall.
\begin{table}[h]
\resizebox{\columnwidth}{!}{
\begin{tabular}{lcccccccc}
\hline
\multirow{2}{*}{\textbf{Method}} & \multicolumn{3}{c}{\textbf{US Politics}} &  & \multicolumn{3}{c}{\textbf{Others}} & \multirow{2}{*}{\textbf{Macro-F1}} \\ \cline{2-4} \cline{6-8}
 & P & R & F1 &  & P & R & F &  \\ \hline
MNLI & 0.925 & 0.428 & 0.585 &  & 0.911 & 0.994 & 0.951 & 0.768 \\
Keywords & \textbf{0.987} & 0.524 & 0.685 &  & 0.925 & 0.999 & 0.961 & 0.823 \\
Hybrid & 0.945 & \textbf{0.710} & \textbf{0.811} &  & 0.953 & 0.993 & 0.973 & \textbf{0.892} \\ \hline
\end{tabular}
}
\caption{CNs data filteration pipeline's quality check on random 1K samples.}
\label{tab:filtered_data_quality}
\vspace{-\baselineskip}
\end{table}

\subsection{Descriptive Statistics of \ds{}}
\label{app:ds_stats}
We present the dataset statistics in Table \ref{tab:desc_stats_data}.
\begin{table*}[!t]
\centering
\resizebox{0.7\textwidth}{!}{
\begin{tabular}{p{0.25\columnwidth}p{0.65\columnwidth}p{0.55\columnwidth}}
\hline
\textbf{Category} & \textbf{Statistic} & \textbf{Value(s)} \\
\hline
\textbf{Overall size} & Number of tweets with notes & 78,698 \\
             & Number of CNs               & 169,992 \\
             & Time span of tweets         & 2021-01-05 -- 2025-09-06 \\
             & Time span of CNs            & 2021-01-28 -- 2025-09-06 \\
             & Language(s)                 & English \\
\hline
\textbf{Contributors} & Unique contributors         & 4,288 \\
             & Median notes per contributor & 1 (IQR: 1--3) \\
             & Max notes by a single contributor & 1,355 \\
\hline
\textbf{Note-level}   & Avg. note length (without URLs)    & 28 words (SD: 13) \\
             & Notes citing external sources      & 77.06\% \\
             & Top sources cited (domains)           & x.com (7.87\%), en.wikipedia.org (2.87\%), apnews.com (2.70\%), en.m.wikipedia.org (2.61\%), www.reuters.com (2.05\%) \\
\hline
\textbf{Tweet-level}  & Unique tweet authors         & 19,537 \\
             & Median follower count of tweet authors & 22,423 \\
             & Median tweet engagement (likes+RTs) & 6216 \\
             & Tweets with multiple notes  & 51\% \\
\hline
\textbf{Helpfulness}  & Notes rated ``Helpful''       & 3.57\% \\
             & Notes rated ``Not Helpful''   & 4.19\% \\
             & Notes ``Needs More Ratings''  & 92.25\% \\
             & Median ratings per note     & 55 (IQR: 15--162) \\
\hline
\textbf{Temporal}     & Median lag: tweet $\rightarrow$ first note & 5H 38M 57S \\
             & Median lag: note $\rightarrow$ status resolution & 5H 31M 57S \\
             & Notes per month (avg.)      & 2,982 \\
\hline
\textbf{Missingness}  & Deleted tweets in dataset    & 12.10\% \\
             & Notes without citation       & 22.93\% \\
\hline
\end{tabular}}
\caption{Descriptive statistics of the \ds{}.}
\label{tab:desc_stats_data}
\end{table*}

\subsection{Computational Resources}
\label{app:comp_res}
All open-models' inference is performed on a commodity GPU cluster (8×NVIDIA H100 GPU cluster) using consistent hyperparameters across models. Models used are publicly available checkpoints (from HuggingFace) or APIs (e.g., OpenAI API).

\subsection{Prompt Templates}
\label{app:prompt_temp}
Figure \ref{fig:sn_lite_prompt} shows the prompt templates for Supernote-Lite, Figure \ref{fig:WA_prompt} for Web-agents, and Figures \ref{fig:GI_prompt}, \ref{fig:TS_prompt}, and \ref{fig:CN_syn_prompt} for the \ours{} framework, covering gap identification, targeted search, and platform-compliant note synthesis, respectively. The prompt used for the LLM-as-a-Judge is provided in Figure~\ref{fig:LLMaaJ_prompt}.
\begin{figure*}[t]
\centering
\begin{tcolorbox}[
    colback=blue!3,
    colframe=blue!35!gray,
    boxrule=0.8pt,
    arc=2pt,
    width=\linewidth,
    title=Supernote Lite
]
\small
You are an expert fact-checker. X (Twitter) has a crowd-sourced fact-checking program, called Community Notes. Here, users can write `notes' on potentially misleading tweets. Each note needs to be rated helpful by a sufficient number of diversely-opinionated people (note-raters) for it to be shown publicly alongside the piece of content.\\

Helpful attributes in notes include:\\
- Cites high-quality sources\\
- Easy to understand\\
- Directly addresses the post's claim\\
- Provides important context\\
- Neutral or unbiased language\\

Unhelpful attributes in notes include:\\
- Sources not included or unreliable\\
- Sources do not support note\\
- Incorrect information\\
- Opinion or speculation\\
- Typos or unclear language\\
- Misses key points or irrelevant\\
- Argumentative or biased language\\
- Note not needed on this post\\
- Harassment or abuse\\

\#\#\# Task\\
You will be given a potentially misleading tweet, its existing community notes, and the helpfulness scores (0-1) of those notes. Your task is to create a single helpful \textbf{Super Community Note} that balances the perspectives reflected in the helpfulness scores while summarizing the key points from the notes. The Supernote should provide clear, factual context that addresses the potentially misleading information in the tweet and be strong enough to replace all existing notes.\\
- The note must be within 280 characters. Treat each URL as 1 character, regardless of its actual length.\\
- The note must include one or more URLs to credible sources. Use only URLs explicitly provided in context. Do not invent or substitute other URLs.\\
- The note must be neutral, factual, and concise. When possible, cite sources across the political spectrum to strengthen neutrality, but prioritize reliability and relevance.\\
- Output only the Super Community Note text, with URLs included. Do not add explanations, formatting, or extra commentary.\\
- Do not include any information beyond what is explicitly provided in the context.\\

--- \\
\textbf{Tweet (published on [date]):}\\
``````[tweet\_text]"""\\

\textbf{Existing Community Note(s):}\\
``````[existing\_notes]"""\\

\textbf{Super Community Note:}
\end{tcolorbox}
\caption{Prompt template for the Supernote-lite baseline. The model is instructed to synthesize a single ``Super Community Note'' by aggregating key points from existing notes, weighted by their helpfulness scores, without performing new external retrieval.}
\label{fig:sn_lite_prompt}
\end{figure*}

\begin{figure*}[t]
\centering
\begin{tcolorbox}[
    colback=violet!3,
    colframe=violet!35!gray,
    boxrule=0.8pt,
    arc=2pt,
    width=\linewidth,
    title=Web Agent
]
\small
You are an expert fact-checker with the ability to use web search tool, enabling you to verify information and write accurate notes to debunk misinformation. X (Twitter) has a crowd-sourced fact-checking program, called Community Notes. Here, users can write `notes' on potentially misleading tweets. Each note needs to be rated helpful by a sufficient number of diversely-opinionated people (note-raters) for it to be shown publicly alongside the piece of content.\\

Helpful attributes in notes include:\\
- Cites high-quality sources\\
- Easy to understand\\
- Directly addresses the post's claim\\
- Provides important context\\
- Neutral or unbiased language\\

Unhelpful attributes in notes include:\\
- Sources not included or unreliable\\
- Sources do not support note\\
- Incorrect information\\
- Opinion or speculation\\
- Typos or unclear language\\
- Misses key points or irrelevant\\
- Argumentative or biased language\\
- Note not needed on this post\\
- Harassment or abuse\\

\#\#\# Task\\
Write a helpful \textbf{Community Note} that clarifies or contextualizes the potentially misleading information in the tweet by providing additional context.\\
- The note must be within 280 characters. Treat each URL as 1 character, regardless of its actual length.\\
- The note must include one or more URLs to credible sources.\\
- The note must be neutral, factual, and concise. When possible, cite sources across the political spectrum to strengthen neutrality, but prioritize reliability and relevance.\\
- Output only the Community Note text, with URLs included. Do not add explanations, formatting, or extra commentary.\\

--- \\
\textbf{Tweet (published on [date]):}\\
``````[tweet\_text]"""\\

\textbf{Community Note:}
\end{tcolorbox}
\caption{Prompt template for the Web-agent baseline. Unlike the gap-driven approach, this agent is given a general instruction to use web search tools to verify the tweet and generate a helpful note in a single stage.}
\label{fig:WA_prompt}
\end{figure*}

\begin{figure*}[t]
\centering
\begin{tcolorbox}[
    colback=green!3,
    colframe=green!40!gray,
    boxrule=0.8pt,
    arc=2pt,
    width=\linewidth,
    title=\ours{}: Gap Identification
]
\small
You are an expert fact-checker and detective analyzing potentially misleading tweets and community notes to find gaps and conflicts.\\

\#\#\# Task\\
Analyze the given potentially misleading tweet and associated community notes (if any) to identify gaps in information, contradictions, or areas needing further investigation. Your goal is to produce a structured list of gaps that require targeted searches to verify facts and provide additional context. Categorize each gap into ONE of these types:\\

UNSUBSTANTIATED\_CLAIM: Factual claims made without sources or evidence\\
Example: ``Studies show X" but no studies are cited\\
If no notes exist: The tweet makes a factual claim that requires verification.\\

CONTRADICTION: Conflicting information\\
Example: Note 1 says ``increases" but Note 2 says ``decreases"\\
If no notes exist: The tweet contains internal contradictions or logical fallacies.\\

VAGUE\_REFERENCE: Non-specific references that should be made concrete\\
Example: ``some studies", ``recent reports"\\

MISSING\_CONTEXT: Statistics, numbers, or claims lacking necessary context\\
Example: ``Crime increased 50\%" without baseline, timeframe, or location\\

SOURCE\_VERIFICATION: Sources mentioned but not provided, or sources need verification\\
Example: ``According to Harvard study" but no link or citation\\

MISSING\_COVERAGE: Important aspects of the tweet not addressed\\
Example: Tweet makes 3 claims but notes only address 1\\
If no notes exist: The entire tweet implies a narrative that lacks context or factual checks.\\

\#\#\# Output Format\\
Return a JSON array of gaps. Each gap should have:\\

gap\_type: One of the 6 types above (exact string match)\\
description: Clear, specific explanation of the gap (1-2 sentences)\\
priority: Integer 1-5, where:\\
5 = Critical (factual claims without sources, major contradictions)\\
4 = High (important context missing, vague references to studies)\\
3 = Medium (minor missing context, secondary claims unsourced)\\
2 = Low (stylistic improvements, minor details)\\
1 = Very low (nice-to-have information)\\
suggested\_query: Specific, targeted search query to fill this gap (be precise)\\

\#\#\# Important Guidelines\\
NO NOTES SCENARIO: If ``EXISTING COMMUNITY NOTES" is empty or ``None", treat every factual claim in the tweet as a potential UNSUBSTANTIATED\_CLAIM or MISSING\_CONTEXT gap needing a search query.\\
Be strategic: Prioritize gaps that most impact credibility and completeness.\\
Be specific: ``Study mentioned but not identified" is better than ``needs more info"\\
Be actionable: Each gap should have a clear, searchable query.\\

---\\
\textbf{Tweet (published on [date]):}\\
``````[tweet\_text]"""\\

\textbf{Existing Community Note(s):}\\
``````[existing\_notes]"""\\

\#\#\# Your Response\\
Analyze the input above and return ONLY the JSON array. No preamble, no explanation, just the JSON.
\end{tcolorbox}
\caption{Prompt template for \ours{} phase I (Gap Identification). The model functions as a detective to analyze the tweet and categorize specific information deficits (e.g., Unsubstantiated Claim, Missing Context) into a structured JSON list of prioritized search queries.}
\label{fig:GI_prompt}
\end{figure*}

\begin{figure*}[t]
\centering
\begin{tcolorbox}[
    colback=green!3,
    colframe=green!40!gray,
    boxrule=0.8pt,
    arc=2pt,
    width=\linewidth,
    title=\ours{}: Targeted Search
]
\small
You are an expert fact-checker with the ability to use web search tool, enabling you to verify information and write accurate fact-checking articles to debunk misinformation.\\

Helpful attributes in fact-checking include:\\
- Cites high-quality sources\\
- Easy to understand\\
- Directly addresses the post's claim\\
- Provides important context\\
- Neutral or unbiased language\\

Unhelpful attributes in fact-checking include:\\
- Sources not included or unreliable\\
- Sources do not support note\\
- Incorrect information\\
- Opinion or speculation\\
- Typos or unclear language\\
- Misses key points or irrelevant\\
- Argumentative or biased language\\
- Note not needed on this post\\
- Harassment or abuse\\

\#\#\# Task\\
Analyze the given potentially misleading tweet and community notes (if any), and use identified information gaps for targeted web searches to retrieve relevant facts from credible sources and synthesize the provided inputs into a short, authoritative fact-checking article. The article should directly address the misleading claims in the tweet, filling the identified gaps with verified information from reliable sources.\\

---\\
\textbf{Tweet (published on [date]):}\\
``````[tweet\_text]"""\\

\textbf{Existing Community Note(s):}\\
``````[existing\_notes]"""\\

\textbf{Identified Gaps and Suggested Queries for Effective Fact-checking:}\\
``````[gap\_context]"""\\

\textbf{Short Fact-checking Article:}
\end{tcolorbox}
\caption{Prompt template for \ours{} phase II (Targeted Search). The agent uses the prioritized gaps from Phase I to guide targeted web retrieval, synthesizing the retrieved evidence into an intermediate, authoritative fact-checking article.}
\label{fig:TS_prompt}
\end{figure*}

\begin{figure*}[t]
\centering
\begin{tcolorbox}[
    colback=green!3,
    colframe=green!40!gray,
    boxrule=0.8pt,
    arc=2pt,
    width=\linewidth,
    title=\ours{}: CN Synthesis
]
\small
You are an expert fact-checker and skilled writer dedicated to producing clear, accurate, and unbiased community notes that debunk misinformation.\\

Helpful attributes in notes include:\\
- Cites high-quality sources\\
- Easy to understand\\
- Directly addresses the post's claim\\
- Provides important context\\
- Neutral or unbiased language\\

Unhelpful attributes in notes include:\\
- Sources not included or unreliable\\
- Sources do not support note\\
- Incorrect information\\
- Opinion or speculation\\
- Typos or unclear language\\
- Misses key points or irrelevant\\
- Argumentative or biased language\\
- Note not needed on this post\\
- Harassment or abuse\\

\#\#\# Task\\
You will be given a potentially misleading tweet and a fact-checking article. Your task is to create a helpful Community Note that balances perspectives reflected in helpfulness and focuses on maximizing the completeness of the note. The note should provide clear, factual context that addresses the potentially misleading information in the tweet and include full-text URLs to support factual information.\\
- The note must be within 280 characters. Treat each URL as 1 character, regardless of its actual length.\\
- The note must include one or more URLs to credible sources.\\
- The note must be neutral, factual, and concise. When possible, cite sources across the political spectrum to strengthen neutrality, but prioritize reliability and relevance.\\
- Output only the Community Note text, with URLs included. Do not add explanations, formatting, or extra commentary.\\
- Do not include any information beyond what is explicitly provided in the context.\\

---\\
\textbf{Tweet (published on [date]):}\\
``````[tweet\_text]"""\\

\textbf{Fact-checking Article:}\\
``````[targeted\_search\_article]"""\\

\textbf{Community Note:}
\end{tcolorbox}
\caption{Prompt template for \ours{} phase III (CN Synthesis). The model is instructed to distill the comprehensive fact-checking article into a concise, platform-compliant note (max 280 characters) that adheres to neutrality and citation standards.}
\label{fig:CN_syn_prompt}
\end{figure*}

\begin{figure*}[t]
\centering
\begin{tcolorbox}[
    colback=yellow!5,
    colframe=yellow!45!gray,
    boxrule=0.8pt,
    arc=2pt,
    width=\linewidth,
    title=LLM-as-a-Judge
]
\small
You are an expert evaluator of community notes. Your task is to score an AI-generated note by comparing it to a human-written helpful note and the original tweet.\\

\#\#\# Evaluation Criteria\\
Functional Errors (1–5): Evaluate whether the AI note has technical or usability issues that reduce its quality, including truncated or incomplete text, broken or incomplete URLs, formatting or punctuation problems, excessive length, or the presence of reasoning or meta-level commentary. A score of 5 means no functional issues at all, while a score of 1 means severe errors that significantly impair usability.\\

Claim Alignment (1–5): Evaluate how accurately the AI note identifies and addresses the primary claim or claims made in the tweet. The note should directly engage with what is actually being asserted or implied, focus on the aspects that require verification, and avoid shifting attention to related but different issues. Proper claim alignment also requires that the facts presented are relevant to the identified claim and help resolve it, rather than merely being topically similar. A score of 5 means the note precisely targets the core claim and supports it with relevant facts, while a score of 1 means the note misunderstands, ignores, or substitutes the main claim with an unrelated or tangential one.\\

Fact Alignment (1–5): Judge whether the factual statements in the AI note are consistent with the human-written note, focusing on whether they describe the same underlying facts, entities, events, timelines, and claims. Different sources are acceptable if they substantively support the same facts, but the alignment must also be verifiable through the cited URLs, including that the source content actually supports the specific statements being made. Topical similarity alone is not sufficient. A score of 5 means all factual claims are fully consistent with the human note and correctly supported by the sources, while a score of 1 means the note contains contradictions, incorrect references, unsupported claims, or factual errors.\\

Completeness (1–5): Evaluate whether the AI note fully covers the key facts and essential context needed to address the tweet’s main claim, as reflected in the human-written note. The note should include all information necessary to understand and verify the claim, without major omissions or reliance on vague or indirect references. Completeness also requires that included content is relevant to the core claim and factually aligned, rather than adding extraneous details. A score of 5 means the note is comprehensive, well-scoped, and factually aligned with the human note, while a score of 1 means critical facts or context related to the claim are missing, misaligned, or insufficient to properly evaluate the claim.\\

Helpfulness (1–5): Evaluate the overall usefulness of the AI note in helping a broad audience understand the post, following community notes standards. A helpful note clearly and directly addresses the post’s claim, uses neutral and unbiased language, and provides important context supported by high-quality, reliable sources that substantiate the stated facts. The note should be easy to understand and focused on information that is actually needed for this post. A score of 5 means the note is clear, well-sourced, relevant, and informative. A score of 1 means the note is unhelpful due to missing or unreliable sources, unsupported or incorrect information, unclear or poorly written text, irrelevant or missing key points, biased or argumentative language, or because the note is unnecessary or inappropriate for the post.\\

\#\#\# Output Format\\
Return ONLY the following JSON structure:\\
\{\\
  "functional\_errors": s1,\\
  "claim\_alignment": s2,\\
  "fact\_alignment": s3,\\
  "completeness": s4,\\
  "helpfulness": s5,\\
  "overall\_comment": "<1–2 lines summarizing overall quality and suitability>"\\
\}\\

---\\
\textbf{Original Tweet (published on [date]):}\\
``````[tweet\_text]"""\\

\textbf{Human-Written Helpful Note (Reference):}\\
``````[human\_note]"""\\

\textbf{AI-Generated Note (To Evaluate):}\\
``````[ai\_note]"""\\

\#\#\# Your Response\\
Analyze the input above and return ONLY the JSON object. No preamble, no explanation, just the JSON.
\end{tcolorbox}
\caption{Prompt template for the LLM-as-a-Judge evaluation. The evaluator is provided with strict rubric definitions to score generated notes on five dimensions: Functional Errors, Claim Alignment, Fact Alignment, Completeness, and Helpfulness.}
\label{fig:LLMaaJ_prompt}
\end{figure*}

\subsection{Model Details}
\label{app:model_details}
The models we used in the experiments are listed in Table \ref{tab:models_overview}, including their model-card names. For open-weight models, we keep $do\_sample=False$ to ensure replicable outputs. We use these models solely for research purposes and in compliance with their permitted use cases.
\begin{table*}[!h]
\centering
\resizebox{0.9\textwidth}{!}
{\begin{tabular}{lllc}
\hline
\textbf{\begin{tabular}[c]{@{}l@{}}Type (Generation\\ Parameters)\end{tabular}} & \textbf{Model Name} & \textbf{Detailed Model Name} & \textbf{\begin{tabular}[c]{@{}c@{}}Tool Used\\ (If Needed)\end{tabular}} \\
\hline
Open-source LLM & Llama-3.1-8B \cite{grattafiori2024llama3} & meta-llama/Llama-3.1-8B-Instruct & -- \\
\multicolumn{1}{c}{(do\_sample=False)} & Ministral-8B \cite{mistralai2024ministral8binstruct} & mistralai/Ministral-8B-Instruct-2410 & -- \\ \hline
Open-source LRM & Apriel-Nemotron-15b \cite{radhakrishna2025apriel} & ServiceNow-AI/Apriel-Nemotron-15b-Thinker & -- \\
\multicolumn{1}{c}{(do\_sample=False)} & Qwen3-14B \cite{yang2025qwen3} & Qwen/Qwen3-14B & -- \\ \hline
Closed-source LRM & GPT-5-nano \cite{singh2026openai} & gpt-5-nano-2025-08-07 & Web search \\
(Default) & Gemini-2.5-Flash \cite{comanici2025gemini} & gemini-2.5-flash & Web search \\
 & Grok-4 \cite{xai2025grok4} & grok-4-0709 & Web search \\
 & Sonar-Deep-Research \cite{perplexity2025sonardeepresearch} & sonar-deep-research & Web search \\ \hline
\end{tabular}}
\caption{An overview of models used in experiments. Note that \textbf{LLM} and \textbf{LRM} are used as short forms for Large Language Model and Large Reasoning Model, respectively.}
\label{tab:models_overview}
\end{table*}

\subsection{Web-Agents based Experiments}
\label{app:web_agent_exp}
Existing Web-Agent systems use different URL citation formats. For example, OpenAI models rely on inline citations, while other agents return source URLs under separate keys in the response JSON. To ensure consistency and minimize information loss, we aggregate all sources into a unified field and append them to the end of the generated note where applicable.

\subsection{Post-processing}
\label{app:post_processing_output}
All AI-generated Community Notes undergo a minor and uniform post-processing step to ensure consistency across models and outputs. This procedure standardizes formatting without altering semantic content by handling null or malformed outputs, trimming extraneous whitespace, removing repeated wrapping quotation marks, and normalizing escaped HTML characters. In addition, model-specific artifacts such as tracking parameters in URLs are stripped to reduce unintended source bias (e.g., \textit{utm\_source=openai} in URL source). Overall, this does not affect the note's quality measures at all.

\subsection{Human Evaluation Guidelines}
\label{app:he_guide}
The questions used in the human evaluation process are shown in Figure \ref{fig:he_guide}. After reviewing the guidelines and task instructions, participants were asked to provide consent for the use of their evaluation data for research purposes.
\begin{figure*}[t]
\centering
\begin{tcolorbox}[
    colback=cyan!5,
    colframe=cyan!40!gray,
    boxrule=0.8pt,
    arc=2pt,
    width=\linewidth,
    title=Guidelines for human evaluation of community notes
]
For a given pair consisting of a potentially misleading tweet and a community note, you need to answer questions covering Factuality, Completeness, and Helpfulness. \\

\textbf{Factuality:} In this part, you need to check if the context mentioned in the note is factually correct and supported by the provided references.

\textbf{F1.} Is the information presented in the note factually correct? (yes/no)

\textbf{F2.} Do the factual sources exist? (yes/no)

\textbf{F3.} Is the information in the note directly supported by the provided sources? (yes/no)

\textbf{F4.} Are the factual sources reliable and credible? (yes/no) \\

\textbf{Completeness:} In this part, you need to evaluate if the community note's short fact-check is complete based on the following categories:

\textbf{C1.} Are the factual claims made in the note relevant to the potentially misleading tweet? (yes/no)

\textbf{C2.} Is the note complete in that it includes all necessary context? (yes/no) \\

\textbf{Helpfulness:} Evaluate how useful this note is in helping a broad audience understand whether the tweet is misleading. Consider whether the note clearly addresses the claim, uses neutral and unbiased language, provides relevant context, and is supported by reliable sources. A highly useful note is clear, informative, and focused on what is needed for this post; a low-usefulness note is unclear, unsupported, irrelevant, biased, or unnecessary. (1-5) \\

\textbf{Ranking:} After evaluating all notes for a given tweet, rank them from most to least useful based on factuality, completeness, and overall helpfulness. The most helpful note should be factually accurate, complete, and clearly explain the context in a neutral and informative way.

\end{tcolorbox}
\caption{The taxonomy of questions presented to annotators. The evaluation task is divided into four categories: Factuality (F1--F4), Completeness (C1--C2), Helpfulness (H), and Ranking (R).}
\label{fig:he_guide}
\end{figure*}

\subsection{\rr{LLM-as-a-Judge Reliability}}
\label{app:llmaaj_human_align}
\rr{We assess agreement between GPT-5.2 Judge and human Helpfulness ratings across 300 jointly evaluated notes. We find a moderate positive association (Spearman $\rho=0.449$; Kendall $\tau=0.376$; both $p<0.001$). This supports using LLM-as-a-Judge for diagnostic evaluation, while the differences between reference-conditioned judging and independent human assessment motivate our reliance on human evaluation for the primary conclusions.}

\rr{\section{Evaluation Details}
\label{app:eval_details}
\subsection{Automated Evaluation}
\label{app:auto_eval}
We report the traditional lexical (ROUGE-L) as well as semantic-similarity-based (BERTScore) metrics and note-level statistics in Table \ref{tab:auto_evals}. Here, the ``\# of Notes'' column presents system success cases out of 488. Additionally, we also discuss the quality checks on the dimension of source credibility and political bias in the forthcoming paragraph.
\begin{table*}[t]
\centering
\resizebox{\textwidth}{!}{
\renewcommand{\arraystretch}{1.2}
\begin{tabular}{c|llccclccccclccccc}
\cline{1-2} \cline{4-6} \cline{8-12} \cline{14-18}
 &  &  & \multicolumn{3}{c}{\textbf{Similarity Scores}} &  & \multicolumn{5}{c}{\textbf{LLM-as-a-Judge}} &  &  &  &  &  &  \\ \cline{4-6} \cline{8-12}
\multirow{-2}{*}{\rotatebox[origin=c]{90}{\textbf{Type}}} & \multirow{-2}{*}{\textbf{Model}} &  & $\bm{R_L}$ $\bm \uparrow$ & \textbf{BS $\bm \uparrow$} & \textbf{URL-R$\bm \uparrow$} &  & \textbf{FE $\bm \uparrow$} & \textbf{CA $\bm \uparrow$} & \textbf{FA $\bm \uparrow$} & \textbf{C $\bm \uparrow$} & \textbf{H $\bm \uparrow$} &  & \multirow{-2}{*}{\textbf{\begin{tabular}[c]{@{}c@{}}Domain\\ Credibility\end{tabular}}} & \multirow{-2}{*}{\textbf{\begin{tabular}[c]{@{}c@{}}Domain\\ Bias\end{tabular}}} & \multirow{-2}{*}{\textbf{\begin{tabular}[c]{@{}c@{}}\# URL\\ Used\end{tabular}}} & \multirow{-2}{*}{\textbf{\begin{tabular}[c]{@{}c@{}}Char\\ Count\end{tabular}}} & \multirow{-2}{*}{\textbf{\begin{tabular}[c]{@{}c@{}} Coverage \\ (\# of Notes)\end{tabular}}} \\ \cline{1-2} \cline{4-6} \cline{8-12} \cline{14-18}
 & Human-written Note &  & --- & --- & --- &  & --- & --- & --- & --- & --- &  & 0.96 & -0.377 & 1.668 & 172 & 488 \\ \cline{1-2} \cline{4-6} \cline{8-12} \cline{14-18}
 & Llama-3.1-8B &  & 0.193 & 0.864 & 0.12 &  & 4.087 & 3.585 & 2.479 & 2.464 & 2.796 &  & 0.935 & -0.55 & 1.808 & 225 & 265 \\
 & Ministral-8B &  & 0.208 & 0.861 & 0.125 &  & 4.083 & 3.589 & 2.487 & 2.355 & 2.672 &  & 0.954 & -0.445 & 1.8 & 154 & 265 \\
 & Qwen3-14B &  & 0.202 & 0.868 & 0.122 &  & 3.974 & 3.728 & 2.615 & 2.487 & 2.838 &  & 0.956 & -0.396 & 1.525 & 147 & 265 \\
 & Apriel-Nemotron-15B &  & 0.192 & 0.862 & 0.107 &  & 3.879 & 3.709 & 2.540 & 2.528 & 2.811 &  & 0.954 & -0.329 & 1.664 & 227 & 265 \\
 & GPT-5-nano &  & 0.17 & 0.856 & \textbf{0.129} &  & 4.053 & 3.668 & 2.634 & 2.506 & 2.860 &  & 0.959 & -0.269 & 1.823 & 212 & 265 \\
\multirow{-6}{*}{\rotatebox[origin=c]{90}{Supernote Lite}} & Gemini-2.5-Flash &  & \textbf{0.224} & \textbf{0.871} & 0.125 &  & 4.358 & \textbf{3.898} & \textbf{2.906} & 2.687 & 3.128 &  & \textbf{0.962} & -0.477 & 1.332 & 161 & 265 \\ \cline{1-2} \cline{4-6} \cline{8-12} \cline{14-18}
 & GPT-5-nano (T1) &  & 0.147 & 0.849 & 0.04 &  & \textbf{4.516} & 3.383 & 2.488 & 2.459 & 2.961 &  & 0.934 & -0.691 & 2.92 & 242 & 488 \\
 & GPT-5-nano (T2) &  & 0.145 & 0.849 & 0.039 &  & 4.496 & 3.340 & 2.465 & 2.426 & 2.984 &  & 0.927 & -0.665 & 2.846 & 242 & 488 \\
 & \cellcolor[HTML]{EFEFEF}$^{\text{GPT-5-nano}}\Delta_{\text{WA}_{T2}-\text{WA}_{T1}}$ &  & \cellcolor[HTML]{EFEFEF}-0.002 & \cellcolor[HTML]{EFEFEF}0 & \cellcolor[HTML]{EFEFEF}-0.001 &  & \cellcolor[HTML]{EFEFEF}-0.02 & \cellcolor[HTML]{EFEFEF}-0.043 & \cellcolor[HTML]{EFEFEF}-0.023 & \cellcolor[HTML]{EFEFEF}-0.033 & \cellcolor[HTML]{EFEFEF}0.023 &  & \cellcolor[HTML]{EFEFEF}-0.007 & \cellcolor[HTML]{EFEFEF}0.026 & \cellcolor[HTML]{EFEFEF}-0.074 & \cellcolor[HTML]{EFEFEF}0 & \cellcolor[HTML]{EFEFEF}0 \\
 & \cellcolor[HTML]{EFEFEF}$^{\text{GPT-5-nano}}\Delta_{\text{WA}_{T2}-\text{SN}}$ &  & \cellcolor[HTML]{EFEFEF}-0.025*** & \cellcolor[HTML]{EFEFEF}-0.007*** & \cellcolor[HTML]{EFEFEF}-0.09*** &  & \cellcolor[HTML]{EFEFEF}0.443*** & \cellcolor[HTML]{EFEFEF}-0.328*** & \cellcolor[HTML]{EFEFEF}-0.169 & \cellcolor[HTML]{EFEFEF}-0.08 & \cellcolor[HTML]{EFEFEF}0.124 &  & \cellcolor[HTML]{EFEFEF}-0.032* & \cellcolor[HTML]{EFEFEF}-0.396*** & \cellcolor[HTML]{EFEFEF}1.023*** & \cellcolor[HTML]{EFEFEF}30*** & \cellcolor[HTML]{EFEFEF}223 \\
 & Gemini-2.5-Flash (T1) &  & 0.171 & 0.857 & 0.043 &  & 4.242 & 3.334 & 2.465 & 2.418 & 2.809 &  & 0.933 & -0.43 & 3.213 & 255 & 488 \\
 & Gemini-2.5-Flash (T2) &  & 0.169 & 0.858 & 0.061 &  & 4.127 & 3.315 & 2.465 & 2.388 & 2.772 &  & 0.943 & -0.416 & 3.884 & 248 & 482 \\
 & \cellcolor[HTML]{EFEFEF}$^{\text{Gemini-2.5-Flash}}\Delta_{\text{WA}_{T2}-\text{WA}_{T1}}$ &  & \cellcolor[HTML]{EFEFEF}-0.002 & \cellcolor[HTML]{EFEFEF}0.001 & \cellcolor[HTML]{EFEFEF}0.018 &  & \cellcolor[HTML]{EFEFEF}-0.115 & \cellcolor[HTML]{EFEFEF}-0.019 & \cellcolor[HTML]{EFEFEF}0 & \cellcolor[HTML]{EFEFEF}-0.03 & \cellcolor[HTML]{EFEFEF}-0.037 &  & \cellcolor[HTML]{EFEFEF}0.01 & \cellcolor[HTML]{EFEFEF}0.014 & \cellcolor[HTML]{EFEFEF}0.671** & \cellcolor[HTML]{EFEFEF}-7 & \cellcolor[HTML]{EFEFEF}-6 \\
 & \cellcolor[HTML]{EFEFEF}$^{\text{Gemini-2.5-Flash}}\Delta_{\text{WA}_{T2}-\text{SN}}$ &  & \cellcolor[HTML]{EFEFEF}-0.055*** & \cellcolor[HTML]{EFEFEF}-0.013*** & \cellcolor[HTML]{EFEFEF}-0.064** &  & \cellcolor[HTML]{EFEFEF}-0.231*** & \cellcolor[HTML]{EFEFEF}-0.583*** & \cellcolor[HTML]{EFEFEF}-0.441*** & \cellcolor[HTML]{EFEFEF}-0.299** & \cellcolor[HTML]{EFEFEF}-0.356*** &  & \cellcolor[HTML]{EFEFEF}-0.019 & \cellcolor[HTML]{EFEFEF}0.061 & \cellcolor[HTML]{EFEFEF}2.552*** & \cellcolor[HTML]{EFEFEF}87*** & \cellcolor[HTML]{EFEFEF}217 \\
 & Grok-4 (T1) &  & 0.167 & 0.857 & 0.114 &  & 1.568 & 3.213 & 2.342 & 2.236 & 1.795 &  & 0.919 & -0.423 & 22.117 & 289 & 488 \\
\multirow{-10}{*}{\rotatebox[origin=c]{90}{Web Agent}} & Sonar-Deep-Res (T1) &  & 0.163 & 0.858 & 0.115 &  & 1.816 & 3.318 & 2.393 & 2.416 & 1.994 &  & 0.945 & -0.364 & 18.84 & 287 & 488 \\ \cline{1-2} \cline{4-6} \cline{8-12} \cline{14-18}
 & GPT-5-nano (T2) &  & 0.145 & 0.848 & 0.043 &  & 4.266 & 3.668 & 2.656 & 2.734 & \textbf{3.141} &  & 0.931 & -0.594 & 3.26 & 245 & 488 \\
 & \cellcolor[HTML]{EFEFEF}$^{\text{GPT-5-nano}}\Delta_{\text{OUR}_{T2}-\text{WA}_{T2}}$ &  & \cellcolor[HTML]{EFEFEF}0 & \cellcolor[HTML]{EFEFEF}{\color[HTML]{FE0000} -0.001} & \cellcolor[HTML]{EFEFEF}{\color[HTML]{009901} 0.004} &  & \cellcolor[HTML]{EFEFEF}{\color[HTML]{FE0000} -0.23***} & \cellcolor[HTML]{EFEFEF}{\color[HTML]{009901} 0.328***} & \cellcolor[HTML]{EFEFEF}{\color[HTML]{009901} 0.191*} & \cellcolor[HTML]{EFEFEF}{\color[HTML]{009901} 0.308***} & \cellcolor[HTML]{EFEFEF}{\color[HTML]{009901} 0.157} &  & \cellcolor[HTML]{EFEFEF}{\color[HTML]{009901} 0.004} & \cellcolor[HTML]{EFEFEF}{\color[HTML]{009901} 0.071} & \cellcolor[HTML]{EFEFEF}0.414*** & \cellcolor[HTML]{EFEFEF}3 & \cellcolor[HTML]{EFEFEF}0 \\
 & \cellcolor[HTML]{EFEFEF}$^{\text{GPT-5-nano}}\Delta_{\text{OUR}_{T2}-\text{SN}}$ &  & \cellcolor[HTML]{EFEFEF}{\color[HTML]{FE0000} -0.025***} & \cellcolor[HTML]{EFEFEF}{\color[HTML]{FE0000} -0.008***} & \cellcolor[HTML]{EFEFEF}{\color[HTML]{FE0000} -0.086***} &  & \cellcolor[HTML]{EFEFEF}{\color[HTML]{009901} 0.213***} & \cellcolor[HTML]{EFEFEF}0 & \cellcolor[HTML]{EFEFEF}{\color[HTML]{009901} 0.022} & \cellcolor[HTML]{EFEFEF}{\color[HTML]{009901} 0.228*} & \cellcolor[HTML]{EFEFEF}{\color[HTML]{009901} 0.281**} &  & \cellcolor[HTML]{EFEFEF}{\color[HTML]{FE0000} -0.028*} & \cellcolor[HTML]{EFEFEF}{\color[HTML]{FE0000} -0.325***} & \cellcolor[HTML]{EFEFEF}1.437*** & \cellcolor[HTML]{EFEFEF}33*** & \cellcolor[HTML]{EFEFEF}223 \\
 & Gemini-2.5-Flash (T2) &  & 0.178 & 0.861 & 0.047 &  & 4.477 & 3.744 & 2.682 & \textbf{2.744} & 3.114 &  & 0.935 & -0.332 & 2.238 & 210 & 484 \\
 & \cellcolor[HTML]{EFEFEF}$^{\text{Gemini-2.5-Flash}}\Delta_{\text{OUR}_{T2}-\text{WA}_{T2}}$ &  & \cellcolor[HTML]{EFEFEF}{\color[HTML]{009901} 0.009} & \cellcolor[HTML]{EFEFEF}{\color[HTML]{009901} 0.003} & \cellcolor[HTML]{EFEFEF}{\color[HTML]{FE0000} -0.014} &  & \cellcolor[HTML]{EFEFEF}{\color[HTML]{009901} 0.35***} & \cellcolor[HTML]{EFEFEF}{\color[HTML]{009901} 0.429***} & \cellcolor[HTML]{EFEFEF}{\color[HTML]{009901} 0.217**} & \cellcolor[HTML]{EFEFEF}{\color[HTML]{009901} 0.356***} & \cellcolor[HTML]{EFEFEF}{\color[HTML]{009901} 0.342***} &  & \cellcolor[HTML]{EFEFEF}{\color[HTML]{FE0000} -0.008} & \cellcolor[HTML]{EFEFEF}{\color[HTML]{009901} 0.084} & \cellcolor[HTML]{EFEFEF}-1.646*** & \cellcolor[HTML]{EFEFEF}-38* & \cellcolor[HTML]{EFEFEF}2 \\
\multirow{-6}{*}{\rotatebox[origin=l]{90}{\ours\ }} & \cellcolor[HTML]{EFEFEF}$^{\text{Gemini-2.5-Flash}}\Delta_{\text{OUR}_{T2}-\text{SN}}$ &  & \cellcolor[HTML]{EFEFEF}{\color[HTML]{FE0000} -0.046***} & \cellcolor[HTML]{EFEFEF}{\color[HTML]{FE0000} -0.01***} & \cellcolor[HTML]{EFEFEF}{\color[HTML]{FE0000} -0.078***} &  & \cellcolor[HTML]{EFEFEF}{\color[HTML]{009901} 0.119*} & \cellcolor[HTML]{EFEFEF}{\color[HTML]{FE0000} -0.154} & \cellcolor[HTML]{EFEFEF}{\color[HTML]{FE0000} -0.224*} & \cellcolor[HTML]{EFEFEF}{\color[HTML]{009901} 0.057} & \cellcolor[HTML]{EFEFEF}{\color[HTML]{FE0000} -0.014} &  & \cellcolor[HTML]{EFEFEF}{\color[HTML]{FE0000} -0.027} & \cellcolor[HTML]{EFEFEF}{\color[HTML]{009901}0.145} & \cellcolor[HTML]{EFEFEF}0.906*** & \cellcolor[HTML]{EFEFEF}49*** & \cellcolor[HTML]{EFEFEF}219 \\ \cline{1-2} \cline{4-6} \cline{8-12} \cline{14-18}
\end{tabular}
}
\caption{\rr{Similarity scores, notes-level statistics, and LLM-as-a-Judge results for AI-generated CNs, using human-written helpful notes as ground truth. 
\textbf{Metrics:} ROUGE-L ($R_L$), BERTScore (BS), and URL Recall (URL-R). 
\textbf{LLM-Judge (1--5):} Functional Errors (FE), Claim Alignment (CA), Fact Alignment (FA), Completeness (C), and Helpfulness (H).
\textbf{Source Quality:} Domain Credibility (0--1) and Domain Bias (-3 to +3).
Gray-shaded rows report differences ($\Delta$); {\color[HTML]{009901}\textbf{green}}/{\color[HTML]{FE0000}\textbf{red}} coloring is shown only for \ours{} comparisons.
Statistical significance (using independent t-test): *$p$$<$$0.05$, **$p$$<$$0.01$, ***$p$$<$$0.001$.}
}
\label{tab:auto_evals}
\vspace{-\baselineskip}
\end{table*}

~
\paragraph{Source Quality and Neutrality.}
We can see in Table \ref{tab:auto_evals} that domain Credibility remains high and comparable across all systems (0.92–0.96), with Human-written notes scoring highest (0.96) and \ours{} close behind (0.931–0.935 across backbones); the modest, occasionally significant drops relative to Supernotes-Lite (e.g., -0.028 for GPT-5-nano, $p<0.05$) reflect the same coverage–fidelity trade-off noted above, as fresh retrieval draws from a broader, less curated pool of sources than existing human notes. On Domain Bias, all systems lean left-of-center on average, consistent with prior findings on partisan skew in cited sources \cite{kuuse2025crowdsourced}; however, \ours{} shows a small numerical shift toward neutrality relative to the corresponding Web-Agent baseline for both backbones, though this difference is not statistically significant, indicating that our closed-book synthesis and gap-conditioned retrieval do not measurably alter the partisan composition of cited sources relative to unconditioned web search.
~
\subsection{Length-Controlled Robustness Check}
\label{app:length_robustness}
We examine whether note length accounts for the helpfulness advantage of our system. \ours{} notes are on average longer than human notes (224.7 vs.\ 187.9 characters), and thus we test whether the helpfulness advantage reduces to verbosity. Regressing helpfulness on note length and system identity (OLS, $N=400$) shows the \ours{} coefficient survives controlling for length ($\beta=0.52$, $p<0.01$), while length itself is not a significant predictor ($p=0.17$). A parallel logistic model of pairwise preference ($N=1{,}200$ ordered comparisons) shows the same pattern. Length difference does not predict which note wins ($p=0.30$), while system identity does ($p<0.001$). Here, Table~\ref{tab:length-ols} reports the helpfulness model; Table~\ref{tab:length-logit} reports the pairwise preference model, built from the same rank comparisons underlying the win-rate matrix in Figure~\ref{fig:he_win_rate}.
~
\begin{table}[!h]
\centering
\small
\begin{tabular}{lrrr}
\toprule
\textbf{Term} & \textbf{Coef.} & \textbf{SE} & \textbf{$p$} \\
\midrule
Intercept          & 3.410   & 0.139 & $<$0.001 \\
GitSearch          & 0.520   & 0.190 & 0.007 \\
SN Lite            & $-$0.228 & 0.190 & 0.232 \\
Web Agent          & 0.109   & 0.192 & 0.569 \\
Length (chars)     & $-$0.0003 & 0.000 & 0.172 \\
\bottomrule
\end{tabular}
\caption{\rr{OLS regression of human helpfulness ($H$) on note length and system identity (reference: Human). $N=400$, $R^2=0.042$.}}
\label{tab:length-ols}
\end{table}

\begin{table}[!h]
\centering
\small
\begin{tabular}{lrrr}
\toprule
\textbf{Term} & \textbf{Coef.} & \textbf{SE} & \textbf{$p$} \\
\midrule
Intercept              & $-$0.260 & 0.117 & 0.026 \\
GitSearch              & 1.093   & 0.171 & $<$0.001 \\
SN Lite                & $-$0.399 & 0.168 & 0.018 \\
Web Agent              & 0.354   & 0.165 & 0.032 \\
Length difference      & $-$0.0001 & 0.000 & 0.296 \\
\bottomrule
\end{tabular}
\caption{\rr{Logistic regression of pairwise preference (row beats column) on length difference and system identity of the row method (reference: Human). $N=1{,}200$ ordered comparisons, pseudo-$R^2=0.051$.}}
\label{tab:length-logit}
\end{table}
~
In both models, note length carries no significant independent association with quality judgments once system identity is accounted for, while the \ours{} coefficient remains positive and significant. This indicates that our method's advantage over human-written notes reflects content quality rather than sheer length.}

\rr{\section{Working Examples}
\label{app:examples}
}
\subsection{Case Study/Error Analysis Examples}
\label{app:case_study_examples}
Table \ref{tab:error_analysis_examples} shows representative samples of misleading post-note pairs using different methods, and their detailed intermediate outputs with observations for \ours{} are given in the following section, Appendix \ref{app:intermediate_outputs}. Here, the misinformation texts are presented as they are on the X platform. 

\subsection{Qualitative Examples of \ours{} Intermediate Outputs}
\label{app:intermediate_outputs}
To address transparency of intermediate outputs, we present \rr{four} end-to-end examples\footnote{\textbf{Note:} Explicit terms are partially censored (e.g.\ \textit{ch*ld p*rn}) to comply with publication guidelines; the pipeline operates on uncensored text.} drawn from our test set.
Each example traces the full pipeline: the input tuple $(t_i, \mathcal{C}_i)$, the structured gap list produced by \textbf{Phase~I}, the evidence article synthesised in \textbf{Phase~II}, and the final platform-compliant note output by \textbf{Phase~III}.
These examples demonstrate how gap detection steers targeted retrieval, and how the retrieved evidence is ultimately distilled into a concise, grounded community note.

\paragraph{Observations.}
Across \rr{Examples~1--3}, Phase~I consistently identifies the most consequential verification barrier first (a \textit{vague reference} gap in \hyperref[example:1]{Example~1}; a \textit{contradiction} about intent in \hyperref[example:2]{Example~2}; dual high-priority \textit{contradictions} in \hyperref[example:3]{Example~3}) and assigns it the highest severity score, directly shaping the retrieval agenda in Phase~II.
\rr{In contrast, \hyperref[example:4]{Example~4} exposes a failure mode. Although Phase~I assigns maximum severity to the \textit{vague reference} involving the unnamed document, Phase~II assumes that ``\emph{this}'' refers to \textit{moharebeh} rather than resolving the referent. This produces a structurally valid but claim-misaligned note (FE$=4$, CA$=1$, FA$=1$, C$=1$, H$=1$; on the scale of 1--5).}
Phase~II's evidence articles are rich, multi-source, and substantially longer than the final notes, confirming that Phase~III is performing genuine distillation rather than mere truncation.
The final notes (Phase~III) are concise, grounded exclusively in retrieved evidence, and include inline citations, consistent with the constrained synthesis design described in Section~\ref{sec:framework}.

\begin{table*}[t]
\centering
\small
\begin{tabular}{p{3.8cm} c p{3cm} c p{6.1cm}}
\toprule
\textbf{Fact-checked Tweet} & \textbf{Rank} & \textbf{Method} & \textbf{H} & \textbf{Note Content (Excerpt)} \\
\midrule

\multirow{4}{4.2cm}{\textit{``Absolutely insane that Democrats actually voted to have the US government publish ch*ld p*rn. WTF is even going on anymore???''}}
&  1 & \textbf{GitSearch (Gemini-2.5-Flash)} & 5 &
Claims that Democrats voted to publish ch*ld p*rn*gr*phy are false. The Epstein Files Transparency Act explicitly prohibits releasing ch*ld s*x**l abuse material and victims’ personal information. \\

& 2 & Human-written CN & 4 &
The Ro Khanna amendment does not compel the government to publish ch*ld p*rn*gr*phy; the sponsor clarified that the DOJ would protect victims’ identities. \\ 

& 3 & Supernotes-Lite (Gemini-2.5-Flash) & 3 &
Publishing ch*ld p*rn*gr*phy is illegal under U.S. law; proposed amendments aim to ensure compliance with existing prohibitions. \\

& 4 & Web-Search Agent (Gemini-2.5-Flash) & 2 &
U.S. law criminalizes the production and distribution of ch*ld p*rn*gr*phy, but does not address the specific legislative claim. \\

\midrule

\multirow{4}{4.2cm}{\textit{``FIRED.''}}
& 1 & \textbf{GitSearch (Gemini-2.5-Flash)} & 4 &
Jerome Powell was not fired on July 16, 2025; President Trump stated that firing him was ``highly unlikely,'' resolving the ambiguity of the original post. \\

& 2 & Supernotes-Lite (Gemini-2.5-Flash) & 3 &
Jerome Powell has not been fired; the post reflects a preference rather than a factual claim. \\

& 3 & Human-written CN & 2 &
Jerome Powell hasn’t been fired; Trump said firing him was ``highly unlikely.'' \\

& 4 & Web-Search Agent (Gemini-2.5-Flash) & 1 &
The tweet lacks sufficient context to interpret the meaning of ``FIRED,'' without resolving the factual status. \\

\midrule

\multirow{4}{4.2cm}{\textit{``I asked Grok to search every record of Trump speaking or writing to determine if he has ever used the word `enigma' before, and Grok says there is no record of him ever saying it.''}}
& 1 & Human-written CN & 5 &
Donald Trump used the word ``enigma'' at least twice in a 2015 rally and in excerpts from his books, contradicting the claim that he never used the word. \\

& 2 & Web-Search Agent (Gemini-2.5-Flash) & 4 &
Trump used ``enigma'' multiple times in rallies and books, citing multiple news sources. \\

& 3 & \textbf{GitSearch (Gemini-2.5-Flash)} & 3 &
Records show Trump used ``enigma'' in speeches and books, contradicting Grok’s claim of no prior usage. \\

& 4 & Supernotes-Lite (Gemini-2.5-Flash) & 3 &
Trump has used the word ``enigma,'' as confirmed by video footage and Grok itself. \\

\bottomrule
\end{tabular}
\caption{Error-analysis examples, with fact-checked tweets shown in the first column. Rows are ordered by human preference (Rank); our method (\ours{}) is in \textbf{bold}. We have not included the URLs' text in the notes for this tabular presentation, and to mask sensitive keywords, we use asterisks (*).}
\label{tab:error_analysis_examples}
\end{table*}


\onecolumn  
\clearpage
\tcbset{
  inputstyle/.style={
    enhanced, breakable,
    colback  = gray!8,
    colframe = gray!55,
    fonttitle= \bfseries\small,
    title    = {Input: Tweet \& Existing \texttt{NMR} Notes},
    left=4pt, right=4pt, top=4pt, bottom=4pt,
  },
  phaseIstyle/.style={
    enhanced, breakable,
    colback  = blue!5,
    colframe = blue!45,
    fonttitle= \bfseries\small,
    title    = {Phase I Output: Detected Information Gaps},
    left=4pt, right=4pt, top=4pt, bottom=4pt,
  },
  phaseIIstyle/.style={
    enhanced, breakable,
    colback  = teal!5,
    colframe = teal!55,
    fonttitle= \bfseries\small,
    title    = {Phase II Output: Targeted Evidence Article},
    left=4pt, right=4pt, top=4pt, bottom=4pt,
  },
  phaseIIIstyle/.style={
    enhanced, breakable,
    colback  = green!6,
    colframe = green!50!black,
    fonttitle= \bfseries\small,
    title    = {Phase III Output: Final Community Note $\hat{n}_i$},
    left=4pt, right=4pt, top=4pt, bottom=4pt,
  },
}

\subsection*{Example 1}
\label{example:1}
\begin{tcolorbox}[inputstyle]
\textbf{Tweet} (Jul 15, 2025):\quad
\textit{``Absolutely insane that Democrats actually voted to have the US government publish ch*ld p*rn WTF Is even going on anymore???''}

\smallskip
\textbf{Existing \texttt{NMR} Notes:}
\begin{enumerate}[leftmargin=*, label=\textbf{N\arabic*.}, itemsep=2pt]
  \item \textit{``They wouldn't release ch*ld p*rn on a government website because it's still illegal to do so. This is just a pathetic excuse to justify not releasing the Epstein files.''} [\url{justice.gov}; \url{merriam-webster.com}]
  \item \textit{``Context matters. Tim said the exact opposite not 6 months ago.''} [\url{x.com/Timcast/...}]
  \item \textit{``NNN — Any records means photos and videos.''}
  \item \textit{``The original amendment would have given a narrow window to allow the publishing of CP hence why it was re-written\ldots''} [\url{twitter.com/timcast/...}]
  \item \textit{``Previous law preventing publishing of explicit material would not prevent the specific effects of a later law. As written, the law would require the release of the videos, likely with victim faces blurred.''} [cites \textit{lex specialis} / \textit{lex posterior}]
\end{enumerate}
\end{tcolorbox}

\begin{tcolorbox}[phaseIstyle]
Five gaps were detected and ranked by severity ($s_g \in \{1,\ldots,5\}$):

\smallskip
\begin{tabularx}{\linewidth}{@{}p{3.2cm}Xc@{}}
\toprule
\textbf{Gap Type} & \textbf{Description} & \textbf{$s_g$} \\
\midrule
\textsc{Vague Reference}
  & The tweet and notes refer to an unspecified vote or amendment.  Identifying the exact bill is prerequisite to all other verification.
  & 5 \\[4pt]
\textsc{Source Verification}
  & Note~4 claims an ``original amendment'' explicitly permitted publication of exploitative material and was later re-written; both versions must be located and compared.
  & 5 \\[4pt]
\textsc{Contradiction}
  & Note~1 asserts illegality bars publication; Note~5 argues a later statute (via \textit{lex posterior}) could override prior law and permit release with redaction. The specific legislative text must adjudicate this.
  & 5 \\[4pt]
\textsc{Unsubstantiated Claim}
  & The tweet asserts Democrats voted for this; the actual roll-call record must be checked once the bill is identified.
  & 4 \\[4pt]
\textsc{Missing Context}
  & Note~1 introduces a motive (deflecting from Epstein files) without evidence. Any legislative link between this bill and Epstein-related materials requires verification.
  & 3 \\
\bottomrule
\end{tabularx}

\smallskip
\textit{High-priority gaps (s$_g \geq 4$)} drove the retrieval queries for Phase~II; the \textsc{Vague Reference} gap was resolved first, with subsequent queries conditioned on the identified legislation (\textit{Epstein Files Transparency Act}).
\end{tcolorbox}

\begin{tcolorbox}[phaseIIstyle]
\textbf{Verdict: False.}

The tweet misrepresents legislative efforts related to the unsealing of Jeffrey Epstein documents.
A House subcommittee voted 8--2 on July~23, 2025 to subpoena the DOJ for the Epstein files\textsuperscript{[1,2]}, reflecting a bipartisan push for \emph{transparency}, not for the publication of exploitative material.

The legislation that ultimately passed --- the \textit{Epstein Files Transparency Act} --- \textbf{explicitly prohibits} the release of ch*ld s*x**l abuse material and victims' personally identifiable information.  Congressional Record H.R.~668 (Sept.~3, 2025) details ``Permitted Withholdings or Redactions,'' which include materials that ``depict or contain ch*ld p*rnography, other ch*ld s*x**l abuse materials, or similar materials''\textsuperscript{[5]}.  PBS NewsHour (Nov.~21, 2025) confirmed that the Act ``bars the release of any materials depicting the s*x**l abuse of ch*ldren''\textsuperscript{[3]}.

Therefore, the votes related to the Epstein files aimed to increase transparency while implementing strict safeguards against the dissemination of exploitative content.  The claim that Democrats voted to publish ch*ld p*rnography is inaccurate and misrepresents both the intent and explicit provisions of the law.

\smallskip\hrule\smallskip
{\footnotesize
\begin{enumerate}[leftmargin=*, label={[\arabic*]}, itemsep=1pt, topsep=0pt]
  \item \url{https://judiciary.house.gov/media/press-releases/judiciary-committee-votes-subpoena-doj-epstein-files}
  \item \url{https://apnews.com/article/epstein-files-house-judiciary-subpoena-doj}
  \item \url{https://www.pbs.org/newshour/politics/7-things-to-know-about-the-justice-departments-epstein-files}
  \item \url{https://www.congress.gov/bill/119th-congress/house-bill/668}
  \item \url{https://www.congress.gov/119/crec/2025/09/03/171/144/CREC-2025-09-03-pt1-PgH3780.pdf}
\end{enumerate}}
\end{tcolorbox}

\begin{tcolorbox}[phaseIIIstyle]
\textit{``Claims Democrats voted to publish ch*ld p*rn are false. The Epstein Files Transparency Act explicitly prohibits releasing ch*ld s*x**l abuse material and victims' PII, while mandating other document releases.''} \\[4pt]
\url{https://www.pbs.org/newshour/politics/7-things-to-know-about-the-justice-departments-epstein-files} \\
\url{https://www.congress.gov/119/crec/2025/09/03/171/144/CREC-2025-09-03-pt1-PgH3780.pdf}
\end{tcolorbox}

\subsection*{Example 2}
\label{example:2}
\begin{tcolorbox}[inputstyle]
\textbf{Tweet} (Jul 16, 2025):\quad \textit{``FIRED.''}

\smallskip
\textbf{Existing \texttt{NMR} Notes:}
\begin{enumerate}[leftmargin=*, label=\textbf{N\arabic*.}, itemsep=2pt]
  \item \textit{``Not fired, no plans (yet).''} [\url{edition.cnn.com/2025/07/16/politics/jerome-powell-donald-trump-firing}]
  \item \textit{``Fake news — Trump said in the interview.''} [\url{x.com/acyn/...}]
  \item \textit{``NNN — In a response to a post stating Jerome Powell must `resign', poster simply expresses a preference for Jerome Powell to be `Fired', and never claims that he is actually `Fired'.''}
\end{enumerate}
\end{tcolorbox}

\begin{tcolorbox}[phaseIstyle]
Three gaps were detected:

\smallskip
\begin{tabularx}{\linewidth}{@{}p{3.2cm}Xc@{}}
\toprule
\textbf{Gap Type} & \textbf{Description} & \textbf{$s_g$} \\
\midrule
\textsc{Contradiction}
  & Notes~1 and~2 treat ``FIRED.'' as a factual claim about Jerome Powell's dismissal, while Note~3 argues it expresses a \emph{preference} in reply to a prior post about resignation. The tweet's original conversational context must be retrieved to resolve the intended meaning.
  & 5 \\[4pt]
\textsc{Source Verification}
  & Note~1 asserts Powell had not been fired; the referenced CNN article must be verified.
  & 4 \\[4pt]
\textsc{Source Verification}
  & Note~2 references a Trump interview via an X post by \texttt{@acyn}; the content must be confirmed to support the ``Fake news'' label.
  & 4 \\
\bottomrule
\end{tabularx}

\smallskip
The high-severity \textsc{Contradiction} gap (resolving \emph{factual claim vs.\ opinion}) was the primary driver of the retrieval strategy, as it determined the entire framing for subsequent verification.
\end{tcolorbox}

\begin{tcolorbox}[phaseIIstyle]
\textbf{Verdict: False (Powell was not fired on July 16, 2025).}

On July~16, 2025, President Trump had publicly criticised Federal Reserve Chair Jerome Powell over interest rates and alleged renovation cost overruns, and had reportedly shown a \emph{draft} termination letter to House Republicans\textsuperscript{[4,5]}. However, Trump publicly clarified that same day that firing Powell was \textbf{``highly unlikely''} unless evidence of ``fraud'' emerged\textsuperscript{[1,2,3]}.

Community Notes~1 and~2 are supported by these public denials\textsuperscript{[2,3,4]}. While Note~3 raised the possibility that the tweet was a statement of preference rather than fact, the widespread circulation of factual claims about Powell's imminent dismissal on that day --- including an explicit post by a Representative stating ``Jerome Powell is going to be fired. Firing is imminent''\textsuperscript{[5]} --- indicates the tweet would foreseeably be read as a factual assertion.

Subsequent days saw Trump visit the Fed (Jul.~24), state he was ``not prepared to fire Jerome Powell'' but reserve the option contingent on rate decisions, and the circulation of a fabricated AI-generated resignation letter further inflamed misinformation\textsuperscript{[9,10,11]}.

\smallskip\hrule\smallskip
{\footnotesize
\begin{enumerate}[leftmargin=*, label={[\arabic*]}, itemsep=1pt, topsep=0pt]
  \item \url{https://www.investopedia.com/trump-denies-planning-to-fire-fed-chair-jerome-powell-11773487}
  \item \url{https://www.reuters.com/world/us/trump-says-no-intention-firing-powell-2025-07-16/}
  \item \url{https://www.washingtonpost.com/video/politics/trump-says-it-is-highly-unlikely-he-will-fire-powell/2025/07/16/7f1c4a4c-7d01-4217-9515-622ad12f5d02_video.html}
  \item \url{https://edition.cnn.com/2025/07/16/politics/jerome-powell-donald-trump-firing}
  \item \url{https://x.com/RepThomasMassie/status/1945528372819042720}
  \item[] \ldots
  \item[{[9]}] \url{https://www.ft.com/content/trump-fed-visit-powell}
  \item[{[10]}] \url{https://www.cnbc.com/2025/07/24/trump-not-prepared-to-fire-powell.html}
  \item[{[11]}] \url{https://apnews.com/article/powell-resignation-letter-ai-fake}
\end{enumerate}}
\end{tcolorbox}

\begin{tcolorbox}[phaseIIIstyle]
\textit{``Jerome Powell was not fired on July 16, 2025. Amid speculation, President Trump publicly clarified that day that firing Powell was `highly unlikely'.''} \\[4pt]
\url{https://www.investopedia.com/trump-denies-planning-to-fire-fed-chair-jerome-powell-11773487} \\
\url{https://www.washingtonpost.com/video/politics/trump-says-it-is-highly-unlikely-he-will-fire-powell/2025/07/16/7f1c4a4c-7d01-4217-9515-622ad12f5d02_video.html}
\end{tcolorbox}

\subsection*{Example 3}
\label{example:3}
\begin{tcolorbox}[inputstyle]
\textbf{Tweet} (Jul 18, 2025):\quad
\textit{``I asked Grok to search every record of Trump speaking or writing to determine if he has ever used the word `cenigma' before, and Grok says there is no record of him ever saying or speaking the word.''}

\smallskip
\textbf{Existing \texttt{NMR} Notes:}
\begin{enumerate}[leftmargin=*, label=\textbf{N\arabic*.}, itemsep=2pt]
  \item \textit{``Reputable search engines (without even doing deep research just quick web search) show Trump used the word enigma.''} [\url{perplexity.ai/...}]
  \item \textit{``There is video footage of Trump using the word `cenigma'.''} [\url{x.com/thiss\_youu/...}]
  \item \textit{``Grok reports correctly that Trump has used the word `cenigma' in the past.''} [\url{x.com/i/grok/...}]
\end{enumerate}
\end{tcolorbox}

\begin{tcolorbox}[phaseIstyle]
Three gaps were detected:

\smallskip
\begin{tabularx}{\linewidth}{@{}p{3.2cm}Xc@{}}
\toprule
\textbf{Gap Type} & \textbf{Description} & \textbf{$s_g$} \\
\midrule
\textsc{Contradiction}
  & The tweet claims Grok found \emph{no} record of Trump using ``enigma.''  Notes~1 and~2 supply direct evidence (search results, video) that he has, directly refuting the tweet's core claim.
  & 5 \\[4pt]
\textsc{Contradiction}
  & The tweet states Grok returned ``no record,'' yet Note~3 links to a Grok share showing the AI \emph{does} find such instances, contradicting the tweet's characterisation of Grok's own output.
  & 5 \\[4pt]
\textsc{Source Verification}
  & The tweet provides no link or screenshot of the alleged Grok query returning ``no record,'' making the foundational premise unverifiable and inherently suspect given Note~3's conflicting result.
  & 4 \\
\bottomrule
\end{tabularx}

\smallskip
Both \textsc{Contradiction} gaps (priority~5) were resolved in parallel during Phase~II retrieval, as each was independently actionable without requiring the other's output first.
\end{tcolorbox}

\begin{tcolorbox}[phaseIIstyle]
\textbf{Verdict: False.}

The claim that Trump has never used the word ``enigma'' is contradicted by public record across multiple media:

\begin{itemize}[leftmargin=*, itemsep=2pt]
  \item \textbf{Public speech:} In November 2015, at a rally in Fort Dodge, Iowa, Trump used the word ``enigma'' \emph{twice} to describe Ben Carson: \textit{``Carson's an enigma to me''} and \textit{``Carson's an enigma.''}\textsuperscript{[1,2,3]} Video footage of this speech is publicly available\textsuperscript{[1,3,4]}.
  \item \textbf{Published books:} ``Enigma'' appears twice in Trump's 2004 book \textit{Trump: How to Get Rich} and once in his 1990 book \textit{Trump: Surviving at the Top}\textsuperscript{[2]}.
  \item \textbf{Alleged correspondence:} A letter attributed to Trump in a 50th birthday book for Jeffrey Epstein reads \textit{``Enigmas never age, have you noticed that?''}\textsuperscript{[1,2,5]} Trump denies authorship, but the other instances remain uncontested.
\end{itemize}

The tweet's claim about Grok's output is also inaccurate: Note~3 directly links to a Grok share in which the AI \emph{does} identify Trump's documented uses of ``enigma.'' This underscores the importance of verifying AI-generated outputs against independently corroborated public records before treating them as authoritative\textsuperscript{[6,7,8]}.

\smallskip\hrule\smallskip
{\footnotesize
\begin{enumerate}[leftmargin=*, label={[\arabic*]}, itemsep=1pt, topsep=0pt]
  \item \url{https://newrepublic.com/post/198123/maga-trump-never-use-one-word-epstein-birthday-letter-enigma}
  \item \url{https://www.motherjones.com/politics/2025/07/wall-street-journal-donald-trump-jeffrey-epstein-birthday-letter-lawsuit/}
  \item \url{https://www.c-span.org/video/?c4554026/user-clip-trump-calls-ben-carson-enigma}
  \item \url{https://www.youtube.com/watch?v=FortDodgeRally2015}
  \item \url{https://archive.org/details/epstein-birthday-book}
  \item \url{https://x.com/i/grok/share/enigma-trump-record}
  \item \url{https://www.factcheck.org/2025/07/grok-trump-enigma/}
  \item \url{https://perplexity.ai/search/trump-enigma-word-usage}
\end{enumerate}}
\end{tcolorbox}

\begin{tcolorbox}[phaseIIIstyle]
\textit{``Records show Donald Trump has used the word `enigma' in public speeches \& books. He used it twice at a 2015 rally describing Ben Carson, and in his books \textit{Trump: How to Get Rich} (2004) \& \textit{Trump: Surviving at the Top} (1990).''} \\[4pt]
\url{https://newrepublic.com/post/198123/maga-trump-never-use-one-word-epstein-birthday-letter-enigma} \\
\url{https://www.motherjones.com/politics/2025/07/wall-street-journal-donald-trump-jeffrey-epstein-birthday-letter-lawsuit/}
\end{tcolorbox}

\subsection*{\rr{Example 4}}
\label{example:4}
\begin{tcolorbox}[inputstyle]
\textbf{Tweet} (Jul 4, 2025):\quad
\textit{``I thought Tucker said this doesn't exist? \\
(And that if it did---bizarrely---that he would nuke Iran?)''}

\smallskip
\textbf{Existing \texttt{NMR} Notes:}
\begin{enumerate}[leftmargin=*, label=\textbf{N\arabic*.}, itemsep=2pt]
  \item \textit{``The document absolutely does indicate that Trump engages in waging war and that his punishment would be death.''} [\url{https://chatgpt.com/share/...}]
  \item \textit{``Under Sharia law, a fatwa is binding to Muslim citizens. `Moharebeh' means resorting to arms to frighten people, and its punishment is exile or death.''} [\url{https://www.un.org/depts/los/LEGISLATIONANDTREATIES/PDFFILES/IRN_national_legislation.pdf}]
\end{enumerate}
\end{tcolorbox}

\begin{tcolorbox}[phaseIstyle]
Six gaps were detected and ranked by severity ($s_g \in \{1,\ldots,5\}$):

\smallskip
\begin{tabularx}{\linewidth}{@{}p{3.2cm}Xc@{}}
\toprule
\textbf{Gap Type} & \textbf{Description} & \textbf{$s_g$} \\
\midrule
\textsc{Source Verification}
  & Note~1 relies on a shared ChatGPT conversation rather than a primary or authoritative document, leaving its central claim without credible substantiation.
  & 5 \\[4pt]
\textsc{Vague Reference}
  & Note~1 refers only to ``the document,'' without giving its title, author, or nature. The document must be identified before its claim about Trump can be verified.
  & 5 \\[4pt]
\textsc{Missing Coverage}
  & The notes do not address the tweet's primary claim that Tucker Carlson said ``this''---presumed by the system to mean \textit{moharebeh} or ``waging war''---does not exist.
  & 4 \\[4pt]
\textsc{Missing Coverage}
  & The notes do not address the tweet's claim that Carlson said he would ``nuke Iran'' if ``it'' existed.
  & 4 \\[4pt]
\textsc{Missing Context}
  & Note~1 introduces Trump, ``waging war,'' and a death penalty without explaining how these claims connect to a tweet about Tucker Carlson.
  & 4 \\[4pt]
\textsc{Missing Context}
  & It is unclear whether Note~1's phrase ``waging war'' denotes the specific Iranian legal concept of \textit{moharebeh}, affecting how the alleged punishment should be interpreted.
  & 3 \\
\bottomrule
\end{tabularx}

\smallskip
The \textsc{Vague Reference} gap was assigned maximum severity. Its suggested query asked: ``What specific document indicates that Trump engages in waging war and that his punishment would be death?''
\end{tcolorbox}

\begin{tcolorbox}[phaseIIstyle]
\textbf{Retrieved conclusion:} there is no credible evidence that Tucker Carlson denied the existence of \textit{moharebeh} or said he would ``nuke Iran'' if it existed.

The retrieval stage interpreted the tweet's pronouns ``this'' and ``it'' as references to \textit{moharebeh}. It retrieved Carlson's broader remarks about Sharia law\textsuperscript{[1]} and his opposition to military action against Iran\textsuperscript{[2]}. It also retrieved legal descriptions of \textit{moharebeh} as an Iranian offence with punishments that can include execution\textsuperscript{[3]}.

For Note~1, the search did not identify a credible document establishing the claim about Trump and instead concluded that the note was unsubstantiated. Crucially, however, it never recovered the specific letter or document denoted by ``this'' and ``the document,'' so the assumed mapping from the pronouns to \textit{moharebeh} remained unverified.

\smallskip\hrule\smallskip
{\footnotesize
\begin{enumerate}[leftmargin=*, label={[\arabic*]}, itemsep=1pt, topsep=0pt]
  \item \url{https://wng.org/opinions/i-dont-think-im-against-that-actually-1762665058}
  \item \url{https://www.mediaite.com/politics/tucker-carlson-warns-trump-against-bombing-iran-will-set-off-a-war-and-result-in-thousands-of-american-deaths/}
  \item \url{https://journalisslp.com/index.php/isslp/article/download/257/581/1915}
\end{enumerate}}
\end{tcolorbox}

\begin{tcolorbox}[phaseIIIstyle]
\textit{``Tucker Carlson has not claimed `Moharebeh' (waging war under Iranian Islamic law) doesn't exist. He has also consistently warned against military action or war with Iran, calling it `suicidal.' Moharebeh is a serious crime in Iran with penalties including execution.''} \\[4pt]
\url{https://wng.org/opinions/i-dont-think-im-against-that-actually-1762665058} \\
\url{https://www.mediaite.com/politics/tucker-carlson-warns-trump-against-bombing-iran-will-set-off-a-war-and-result-in-thousands-of-american-deaths/} \\
\url{https://journalisslp.com/index.php/isslp/article/download/257/581/1915}
\end{tcolorbox}

\end{document}